\RequirePackage{fix-cm}
\documentclass[smallextended]{svjour3}
\smartqed
\usepackage{graphicx}
\usepackage{amsmath}
\usepackage{amssymb} 
\usepackage{xcolor}
\usepackage{colortbl}   
\usepackage{tabularx}
\usepackage{xltabular}  
\usepackage{array}
\usepackage{microtype}
\usepackage{lmodern}       
\usepackage[T1]{fontenc}   
\usepackage[utf8]{inputenc}
\usepackage{comment}

\usepackage{booktabs}
\newenvironment{scheme}
  {\par\addvspace{6pt}%
   \setlength{\aboverulesep}{4pt}%
   \setlength{\belowrulesep}{6pt}%
   \noindent\hspace*{3.5em}%
   \begin{tabular}{@{}r@{\hspace{0.9em}}l@{}}}
  {\end{tabular}\par\addvspace{6pt}}

\usepackage[american]{babel}  
\usepackage{csquotes}         
\usepackage{bm}

\definecolor{g75}{gray}{0.35}
\definecolor{g50}{gray}{0.55}
\definecolor{g28}{gray}{0.75}
\definecolor{g12}{gray}{0.90}

\newcommand{\agreebar}[4]{%
  \begin{tikzpicture}[baseline=-0.5ex]
    \def\bw{7}
    \def\bh{7pt}
    \pgfmathsetmacro{\wa}{#1/100*\bw}
    \pgfmathsetmacro{\wb}{#2/100*\bw}
    \pgfmathsetmacro{\wc}{#3/100*\bw}
    \pgfmathsetmacro{\wsegd}{#4/100*\bw}
    \pgfmathsetmacro{\xb}{\wa}
    \pgfmathsetmacro{\xc}{\wa+\wb}
    \pgfmathsetmacro{\xd}{\wa+\wb+\wc}
    \fill[g75] (0,0)       rectangle (\wa cm, \bh);
    \fill[g50] (\xb cm, 0) rectangle (\xc cm, \bh);
    \fill[g28] (\xc cm, 0) rectangle (\xd cm, \bh);
    \fill[g12] (\xd cm, 0) rectangle (\bw cm, \bh);
    \pattern[pattern=north east lines, pattern color=g75]
      (\xb cm, 0) rectangle (\xc cm, \bh);
    \pattern[pattern=north west lines, pattern color=g50]
      (\xc cm, 0) rectangle (\xd cm, \bh);
    \pattern[pattern=crosshatch, pattern color=g28]
      (\xd cm, 0) rectangle (\bw cm, \bh);
    \draw[white, line width=0.7pt] (\xb cm, 0) -- (\xb cm, \bh);
    \draw[white, line width=0.7pt] (\xc cm, 0) -- (\xc cm, \bh);
    \draw[white, line width=0.7pt] (\xd cm, 0) -- (\xd cm, \bh);
    \draw[gray!40, line width=0.3pt] (0,0) rectangle (\bw cm, \bh);
    \pgfmathsetmacro{\mida}{\wa/2}
    \pgfmathsetmacro{\midb}{\xb+\wb/2}
    \pgfmathsetmacro{\midc}{\xc+\wc/2}
    \pgfmathsetmacro{\midd}{\xd+\wsegd/2}
    \pgfmathparse{#1>3 ? 1 : 0}\ifnum\pgfmathresult=1
      \node[font=\footnotesize, below=-1pt] at (\mida cm, 0) {#1\%};
    \fi
    \pgfmathparse{#2>3 ? 1 : 0}\ifnum\pgfmathresult=1
      \node[font=\footnotesize, below=-1pt] at (\midb cm, 0) {#2\%};
    \fi
    \pgfmathparse{#3>3 ? 1 : 0}\ifnum\pgfmathresult=1
      \node[font=\footnotesize, below=-1pt] at (\midc cm, 0) {#3\%};
    \fi
    \pgfmathparse{#4>3 ? 1 : 0}\ifnum\pgfmathresult=1
      \node[font=\footnotesize, below=-1pt] at (\midd cm, 0) {#4\%};
    \fi
  \end{tikzpicture}%
}

\newcommand{\agreekeyplain}[1]{%
  \begin{tikzpicture}[baseline=-0.15ex]
    \fill[#1] (0,0) rectangle (0.8em,0.8em);
    \draw[gray!50, line width=0.3pt] (0,0) rectangle (0.8em,0.8em);
  \end{tikzpicture}%
}
\newcommand{\agreekey}[3]{%
  \begin{tikzpicture}[baseline=-0.15ex]
    \fill[#1] (0,0) rectangle (0.8em,0.8em);
    \pattern[pattern=#2, pattern color=#3] (0,0) rectangle (0.8em,0.8em);
    \draw[gray!50, line width=0.3pt] (0,0) rectangle (0.8em,0.8em);
  \end{tikzpicture}%
}

\newcommand{\dimkeyrecon}{%
  \begin{tikzpicture}[baseline=-0.55ex]
    \draw[black, line width=1.1pt] (0,0) -- (0.8,0);
    \fill[black] (0.4,0) circle (2.1pt);
  \end{tikzpicture}}
\newcommand{\dimkeysurf}{%
  \begin{tikzpicture}[baseline=-0.55ex]
    \draw[g75, line width=1.0pt, dash pattern=on 3.6pt off 2.2pt] (0,0) -- (0.8,0);
    \draw[g75, line width=0.85pt, fill=white]
      (0.4,0) ++(-2.3pt,-2.3pt) rectangle ++(4.6pt,4.6pt);
  \end{tikzpicture}}
\newcommand{\dimkeyunmeas}{%
  \begin{tikzpicture}[baseline=-0.55ex]
    \draw[black, line width=1.1pt, line cap=round,
          dash pattern=on 0pt off 3.2pt] (0,0) -- (0.8,0);
    \draw[black, line width=0.9pt, fill=white] (0.4,0) circle (2.4pt);
  \end{tikzpicture}}

\definecolor{splitprim}{gray}{0.13}
\definecolor{splitcross}{gray}{0.42}
\definecolor{splitext}{gray}{0.78}
\newcommand{\spPrim}[1]{\cellcolor{splitprim}\textcolor{white}{\textbf{#1}}}
\newcommand{\spCross}[1]{\cellcolor{splitcross}\textcolor{white}{\textbf{#1}}}
\newcommand{\spExt}[1]{\cellcolor{splitext}\textbf{#1}}
\newcommand{\spNone}{\textcolor{black!45}{---}}
\newcommand{\spKey}[1]{\textcolor{#1}{\rule[-0.05em]{0.75em}{0.75em}}}

\newcommand{\repoURL}{\url{https://github.com/metabolean5/enthymeme-resource}}

\newcommand{\datasetLicence}{\href{https://creativecommons.org/licenses/by/4.0/}%
{Creative Commons Attribution 4.0 International (CC BY 4.0)}}

\usepackage[htt]{hyphenat}

\usepackage{tikz}
\usetikzlibrary{arrows.meta, positioning, fit, backgrounds, patterns, calc}

\usepackage[hidelinks]{hyperref}
\usepackage{orcidlink}   

\usepackage[style=apa, natbib=true, backend=biber, uniquename=false]{biblatex}
\DeclareLanguageMapping{american}{american-apa}
\journalname{Lang Resources \& Evaluation}

\begin{document}

\title{A Resource for Enthymeme Detection in Controversial Political Discourse}

\author{Martial Pastor \and Nelleke Oostdijk}

\institute{
    Martial Pastor (corresponding author) \at
    Centre for Language Studies, Radboud University, Nijmegen \\
    ORCID: \orcidlink{0009-0004-8128-5497}~\href{https://orcid.org/0009-0004-8128-5497}{0009-0004-8128-5497} \\
    \email{martial.pastor@ru.nl}
    \and
    Nelleke Oostdijk \at
    Centre for Language Studies, Radboud University, Nijmegen \\
    ORCID: 
    \orcidlink{0000-0002-4858-7547}~\href{https://orcid.org/0000-0002-4858-7547} {0000-0002-4858-7547}\\
    \email{nelleke.oostdijk@ru.nl}
}

\date{Received: date / Accepted: date}
\maketitle

\begin{abstract}
Enthymemes — arguments with unstated premises or conclusions — are pervasive in persuasive discourse, yet their annotation remains notoriously subjective. We present a resource of 1,482 tweets from politically controversial discourse, annotated for the presence of enthymemes and their argument structure by five annotators and two validators, with each tweet receiving at least three independent annotations and two validations, designed to study label variation. We first revisit the definition of enthymemes and propose annotation guidelines anchored in Walton's argumentation schemes, offering a structured and constrained approach that nonetheless preserves room for the interpretive nature of the task. This contrasts with past resources, which tend to eliminate disagreement, obscuring its sources and preventing investigation of its potential benefits for NLP model development. We further propose a complexity analysis of the task, identifying where annotation imposes high cognitive load and may give rise to inconsistent annotation. Our preliminary experiments show that models trained on annotator disagreement can outperform models trained on hard majority-vote labels. We close by reflecting on how structural openness in enthymeme definitions and guidelines enables the study of variation in subjective inferential processes for future resources and downstream NLP applications concerned with human inference.
\end{abstract}

\section{Introduction}
\label{sec:intro}

This paper presents a dataset of enthymemes — arguments with missing premises or conclusions — built to study implicit argumentation in controversial political discourse. Though enthymemes occur naturally in everyday discourse, they can obscure questionable assumptions or controversial content, and are sometimes weaponized as persuasive tools to disseminate misinformation. To support argumentation mining research and broader Natural Language Processing (NLP) and other computational experiments, we introduce a resource of 1,482 annotated tweets labeled for the presence or absence of enthymemes and, when present, for their argument structure.

The resource differs from prior work in two respects. First, it restricts annotation to implicit content that expresses a stance on one of two controversial political topics (covid and immigration), directly engaging with the persuasive and potentially manipulative function of enthymemes in political discourse. Second, it is grounded in a formal definition of enthymematic arguments that draws on Walton's argumentation schemes \citep{walton2008}: annotators must identify the inferential pattern underlying an argument, reify its license as a propositional conditional, and reconstruct the missing premise or conclusion such that classical entailment is restored. This definition serves two purposes simultaneously — the entailment requirement makes the completeness of every reconstruction structurally verifiable, while the scheme requirement constrains its content, preserving the interpretive space within which genuine annotator variation can legitimately occur.

The remainder of the paper is organized as follows. Section~\ref{sec:background} situates the work within argumentation theory, surveys existing enthymeme resources and outlines how our work diverges from them. Section~\ref{sec:theory} introduces the theoretical framework, grounding our definition of enthymematic arguments in Walton's argumentation schemes and classical propositional logic. Section~\ref{sec:data} describes the data source and resource statistics. Section~\ref{subsec:annotation-framework-development} details how the annotation guidelines were developed. Section~\ref{sec:dimensions} presents a complexity analysis of the annotation task across the six dimensions of \citet{fort-etal-2012-modeling}, documenting the cognitive load involved and establishing why moderate inter-annotator agreement is structurally expected in tasks like the present one requiring semantic interpretation. Section~\ref{sec:guidelines} summarizes the resulting annotation guidelines. Sections~\ref{sec:process} and~\ref{sec:statistics} report on the four-month annotation campaign involving five annotators and two validators and on the final dataset statistics. The preliminary computational experiments in Section~\ref{sec:experiments} then show that annotator disagreement, preserved through soft-label training, yields gains over majority-vote baselines, with the most pronounced improvements precisely on the instances annotators found hardest to agree on. Section~\ref{sec:discussion} returns to the nature of that disagreement, distinguishing inconsistencies driven by task complexity from genuine disagreement rooted in subjective inferential processes, and tracing these two types to the semantic and pragmatic levels respectively.

\noindent The main contributions of this paper are as follows:

\begin{itemize}
    \item A definition of enthymemes for annotation and computational 
    analysis, grounded in classical propositional logic and Walton's 
    argumentation schemes \citep{walton2008}, in which argumentation schemes
    constrain the content of reconstructed premises and conclusions, and
    classical entailment serves as a mechanically verifiable completeness check.
    
    \item A resource of 1,482 annotated tweets, each labeled by at least three
    annotators and reviewed by two validators, for the presence or
    absence of enthymemes and, when present, for their argument structure
    and propositional content segments.
    
    \item A complexity analysis of the enthymeme annotation task following 
    the six dimensions of \citet{fort-etal-2012-modeling}, documenting the 
    cognitive load of the task and establishing why moderate inter-annotator agreement is expected in similar tasks requiring semantic interpretation.
    
    \item Preliminary computational experiments showing that annotator
    disagreement, preserved through soft-label training, yields gains in enthymeme
    detection over majority-vote baselines.
    
    \item A discussion of annotator disagreement in enthymeme annotation, 
    distinguishing inconsistent annotations arising from task complexity from 
    genuine disagreement reflecting irreducible differences in annotator 
    perception and inferential reconstruction of implicit meaning.
\end{itemize}

\section{Background and Related Work}
\label{sec:background}

\subsection{Implicit Argumentation and Manipulation}
\label{sec:background-implicit}

The use of enthymemes in natural language is caught between communicative economy and rhetorical persuasion.

When a premise is trivial or mutually accepted, making it explicit would only introduce redundancy and slow discourse down. In this sense, enthymemes are a natural linguistic phenomenon, fully compliant with and arguably emergent from Grice's Maxim of Quantity \citep{grice1975}: \textit{do not make your contribution more informative than required}. Enthymemes should thus be understood as obeying a pragmatic principle that discourages speakers from stating what the audience already accepts \citep{Razuvayevskaya2017}. They are accordingly very frequent in everyday discourse, and speakers and hearers alike produce and process implicit content with little to no cognitive effort.

It is precisely this prevalence and cognitive ease that allow defeasible, doubtful, or contentious propositions to enter discourse imperceptibly, slipping through, as it were, under the cover of what feels like routine inferential processing. Indeed as \citet{walton2005argumentation} point out, most implicit premises are often defeasible as speakers prefer not to state potentially weak reasons explicitly. This reticence, \citet{reboul2011relevance} argues, is rarely innocent: implicit communication frequently serves to hide manipulative intentions. That said, implicit content still manages to carry whole discourses forward, resting on a foundation of clay.

Considering these manipulative intentions and potential weaponization of enthymemes, several scholars in linguistics and argumentation research have thus investigated how implicit content in political propaganda can introduce contestable propositions past critical scrutiny without triggering resistance. \citet{lombardivallauri2020implicit} show in an analysis of Italian political campaign slogans (2006) that contestable propositions conveyed via implicit content (such as the suggestion that the Left would introduce property taxes or open borders) went largely unchallenged by voters, whereas the same claims stated explicitly in pilot versions provoked immediate resistance and were deemed implausible. In a different study, \citet{macagno2022argumentation} shows that political leaders make systematic use of unstated premises on Twitter to build tacit consensus. Through corpus linguistic analysis of tweets by four political leaders (Salvini, Trump, Bolsonaro, Biden) over six months, he demonstrates how unshared presuppositions are systematically used so as to present contested claims as already part of the common ground, to build tacit consensus intelligible only to those already sharing their political worldview, effectively bypassing critical evaluation. 

According to \citet{reboul2011relevance}, the persuasive force of enthymemes derives from the dynamic whereby when a statement is left implicit, readers must reconstruct it themselves, leading them to perceive the unstated message as their own and therefore to believe it more readily. The resource presented here illustrates how Twitter instantiates this dynamic, rendering online discourse a particularly fertile ground for the propagation of contentious and potentially false claims, while also filling what is, to the best of our knowledge, a gap in resources for systematically tracking such enthymemes through computational methods. As \citet{stalnaker2002} observes, common ground accumulates incrementally through communicative exchange, meaning that implicitly introduced premises, left unchallenged, quietly sediment into shared belief, a process that Twitter may accelerate and scale, fostering bubbles in which misinformation circulates beneath the threshold of explicit argumentation \citep{nguyen2020}.

\subsection{Enthymemes in Argumentation Theory}
\label{sec:background-theory}

Originating from the rhetorical traditions of the ancient world, the term \textit{enthymeme} first appears in Aristotle's \textit{Prior Analytics}, where it is said to refer to an incomplete (\textit{ateles}) syllogism (\textit{sullogismos}). The authenticity of this definition has since been questioned by \citet{burnyeat1994}, who argues that the relevant passage may have been altered by early commentators and that Aristotle's own conception was more rhetorical than logical in nature, that is, grounded in doxa-based, audience-directed argument rather than formal deduction\footnote{As evidenced by the treatment found in the \textit{Rhetoric} and the \textit{Topics}.} \citep{walton2008}. Nevertheless, the idea of a standard-form syllogism with a missing proposition, a truncated syllogism, proved durable and persisted well into modern logic, where enthymemes have been approached through their deductive facet \citep{govier1992}.\footnote{This deductivist tradition, however, faces a well-known objection: for any argument whatsoever, deductive validity can be restored trivially by adding as a premise the conditional linking the stated premises to the conclusion, so that validity alone cannot distinguish a faithful reconstruction from a vacuous one \citep{hitchcock1985,govier1992}.} In recent developments in argumentation theory, the rhetorical and pragmatic reality has somewhat caught up with the term. Purely syllogistic conceptions of the enthymeme have received less attention \citep{hitchcock1985}, giving way to a conception of the enthymeme as a form of reasoning that holds unless there is evidence to the contrary, grounded in defeasible patterns of inference \citep{walton2008}.

Indeed, syllogistic logic does not apply well to the reconstruction of missing premises or conclusions in everyday language use, where people rely on unstructured sets of reasons to support their claims rather than on predefined propositional structures \citep{pollock1991}. But even beyond this formal inadequacy, attempts at formalizing defeasible patterns of inference face a more fundamental impasse. As \citet{walton2008} state, attributing unstated assumptions to an arguer is an inherently perilous endeavor: even with the most sophisticated formalism, reconstruction ultimately requires interpreting what an arguer meant to say, exposing the analyst to the charge of committing a \textit{straw man fallacy}, that is, distorting or exaggerating an opponent's argument in ways that make it more vulnerable to refutation than it actually is.

However difficult or even unfeasible the task appears, any plausible attempt at enthymeme reconstruction would require both a thorough analysis of argumentative structure and a careful engagement with the domain-specific knowledge in which a given enthymeme is uttered. \citet{sviridova2025} survey a range of argumentation models that offer flexible frameworks for this purpose, spanning approaches that prioritize rhetorical effectiveness and audience agreement \citep{perelman1991}, those that articulate rebuttal strategies suited to the defeasible nature of inference \citep{freeman2011}, or those that analyze the pragmatic functions of argumentation within dialogue \citep{vaneemeren2004}. Among these, Walton's argumentation schemes have proved most influential, capturing stereotypical patterns of what he terms presumptive reasoning that are defeasible by design \citep{walton2008}. It is this theoretical framework that we draw upon for the definition of enthymeme we put forward here, both in our annotation guidelines and in our computational approach.

\subsection{Existing Argument Mining Resources for Enthymemes}
\label{sec:background-resources}

Work on the development of implicit content datasets (and enthymemes datasets in
particular) spans a wide range of disciplines and methodologies. We review in
Table~\ref{tab:resources} the research most closely related to our resource, comparing
each along the axes on which our approach diverges: the domain annotated, the structural
constraint the annotation scheme imposes on the reconstruction, whether the annotation
captures the annotator's interpretation at all, and how agreement is reported.

\begin{table}[ht]
\centering
\caption{Existing resources for implicit argumentation and enthymemes, compared with the
present work, ordered by the strength of the structural constraint imposed on the
reconstruction. Agreement figures are those reported by each original paper; where the
resource reports agreement only for an auxiliary step, this is stated}
\label{tab:resources}
\footnotesize
\setlength{\tabcolsep}{3pt}
\renewcommand{\arraystretch}{1.15}
\newcolumntype{L}{>{\raggedright\arraybackslash}X}
\begin{tabularx}{\textwidth}{@{}>{\raggedright\arraybackslash}p{1.6cm}
                             >{\raggedright\arraybackslash}p{1.45cm}
                             L L L
                             >{\raggedright\arraybackslash}p{2.1cm}@{}}
\toprule
\textbf{Resource} & \textbf{Domain} & \textbf{Size} & \textbf{Structural constraint} &
\textbf{Interpretation captured} & \textbf{Agreement} \\
\midrule
\citet{wojatzki2016stance}
  & Tweets, atheism
  & 733 tweets
  & Stance label only; no argument structure
  & None; target stances known in advance
  & Fleiss $\kappa$ 0.63 (joint stance decision) \\
\citet{stahl2023mind}
  & Learner essays (ICLEv3)
  & 40,089 instances
  & Argumentative discourse units deleted automatically
  & None; no human interpretive step
  & Not applicable \\
\citet{boltuzic2016fill}
  & Online debates, four topics
  & 500 claim pairs
  & Free-text premise linking to a predefined conclusion
  & Reconstruction, otherwise unconstrained
  & None; 32\% word overlap between annotators \\
ARCT \citep{habernal2018argument}
  & Debate claims
  & 1,970 instances
  & One missing warrant, with a plausible alternative
  & Free crowd reconstruction; inference chain untracked
  & Cohen's $\kappa$ 0.58 stance, Krippendorff's $\alpha_u$ 0.51 spans; none for warrants \\
\citet{saha2021irac}
  & Debatable propositions, 71 debate topics
  & 3,166 pairs
  & One defeasible causal link (action to outcome)
  & Causal inference only
  & Fleiss $\kappa$ 0.61 (stance); none for graphs \\
\citet{becker2020implicit}
  & Expert-constructed short texts
  & 719 unit pairs
  & Minimal sentence set making the link explicit
  & Full explicit reconstruction
  & Semantic similarity, not IAA \\
\midrule
This work
  & Tweets, Immigration and COVID-19
  & 1,482 tweets
  & Propositions in premise\slash conclusion slots, plus argumentation scheme
  & Full explicit reconstruction, 3--5 annotators per item
  & Krippendorff's $\alpha = 0.516$ (2-class) \\
\bottomrule
\end{tabularx}
\end{table}

Two gaps emerge. The first concerns interpretation. Where the annotation imposes no
structural constraint, annotators do not engage with the interpretive nature of
enthymemes; where the data is generated by deletion, the resulting text, though it
contains an implicit argumentative unit, is not a naturally occurring enthymeme at all.
Where the reconstruction is constrained, it is constrained narrowly---to a single warrant
that is a creative approximation of the missing link rather than a complete and
verifiable reconstruction of the inference chain, or to a single causal link whose
implicit components often articulate neutral background knowledge rather than politically
oriented propositions.

The second concerns disagreement. \citet{becker2020implicit} is the closest in spirit to
ours in its commitment to making implicit content fully explicit through natural language
reconstruction, but remains limited to a small corpus of carefully constructed expert
arguments and measures agreement through semantic similarity rather than traditional
inter-annotator agreement. How annotators diverge in their interpretation of what is
missing therefore remains open---the question our resource addresses, in a noisier and
more `in the wild' setting, by requiring that all propositions necessary for the logical
entailment to hold be made explicit and by retaining those divergent interpretations
rather than resolving them away.


\section{Theoretical Framework}
\label{sec:theory}

Our approach to enthymeme reconstruction draws on Walton's theory of argumentation schemes \citep{walton2008}. Argumentation schemes are stereotypical patterns of defeasible reasoning, such as \textit{argument from authority} or \textit{argument from cause to effect}, that capture the inferential licenses routinely exploited in natural argumentation. Adopting this framework gives annotators a constrained basis for reconstruction: rather than freely hypothesizing any implicit content, annotators must identify a scheme that licenses the inferential step and use it to motivate what was left unsaid. The scheme both guides and constrains the reconstruction.

This design responds to a shared limitation across existing resources: none requires the reconstructed content to be both structurally complete and tied to an explicit inferential pattern. Approaches in the literature either impose no structural constraint on reconstruction \citep{wojatzki2016stance,boltuzic2016fill}, bypass interpretation entirely \citep{stahl2023mind}, allow freely written warrants that may rest on further unstated assumptions \citep{habernal2018argument,becker2020implicit}, or reduce implicit content to a single causal link \citep{saha2021irac}. We diverge from all of the above approaches by requiring that the defeasible reasoning be captured through a specific argumentation scheme whose inferential license is reified as a propositional conditional and injected into the premise set. Reconstruction is complete when the explicit premises together with all recovered implicit content --- missing premises, the reified scheme, and where necessary an implicit conclusion --- classically entail the conclusion. In this design, entailment functions as a completeness check --- it guarantees that no step of the inference remains hidden --- while the scheme annotation does the substantive work of constraining \textit{which} conditional may be reconstructed, and preserves the defeasible character of the original reasoning. The formal apparatus through which this is achieved is described in Section~\ref{sec:formal}.

\subsection{Argumentation Schemes}
\label{sec:schemes}

Argumentation schemes emerged in the 1960s as part of a broader effort to characterize the inferential principles underlying defeasible reasoning \citep{hastings1963,vaneemeren1978}. They capture stereotypical patterns of argumentation that are conventionally accepted in argumentative communication, re-establishing the inferential link between the premises and the conclusion when that link is left implicit \citep{Visser2022}. The concept has since been approached through a variety of classifications, and has found application in symbolic artificial intelligence and computer science, where schemes have been used to support automated reasoning systems \citep{aracuria, walton2008, macagno2021}.

Walton's taxonomy has proved the most influential, with ongoing work extending it to capture a great variety of schemes in detail \citep{Visser2022}. At the same time, its classification is flexible enough to allow adjustments for domain-specific applications and research projects \citep{walton2008, atkinson2018, kokciyan2018}.

An enthymeme in our resource can only exist if an argument is being made, and any argument necessarily relies on some form of inferential pattern operating in the background. Identifying this pattern is therefore central to enthymeme reconstruction. The flexibility of Walton's taxonomy is particularly suited to this task: not all authors in social media discourse rely on formally valid schemes, and some arguments exploit fallacies which, while shallow and easily defeasible, can be reconstructed within the same scheme apparatus --- their persuasive force deriving precisely from their resemblance to legitimate inferential patterns \citep{walton2011}. Given our focus on recovering implicit meaning rather than evaluating argumentative validity, the goal is not to select a formally correct scheme label or diagnose a fallacy, but simply to reconstruct the type of inferential pattern the author is exploiting to connect premises to their conclusion.

Two of the most frequently occurring inferential patterns in the dataset are 
illustrated below. Each scheme is presented in its abstract form, followed 
by a representative instance drawn from the dataset. Consider the following example:

\begin{quote}
\textit{``But you can mandate a fucking useless vaccine and strip me of my 
rights??? you should go to jail.''} \#1845
\end{quote}
\noindent The scheme \textit{Argument from moral judgment} takes the following form:

\begin{scheme}
$P_1$: & Action $A$ is morally wrong \\[2pt]
$P_2$: & Agent $X$ performed action $A$ \\
\midrule
$C$:   & Therefore, agent $X$ acted wrongly and should be condemned
\end{scheme}

\noindent In this example, $P_2$ is explicit, the speaker asserts that the addressee mandated a useless vaccine and stripped them of their rights. The conclusion $C$ is also explicit, the person addressed should be condemned, while $P_1$, that such an action is morally wrong, is left implicit. If the annotator can identify the operative argument scheme and demonstrate its instantiation in the tweet, they may annotate the presence of an implicit premise. Consider the following example:

\begin{quote}
\textit{``Where was `the right to make the best decisions for your health' 
when your administration was compelling pregnant women to take experimental 
vaccines, hypocrite?''} \#2030
\end{quote}

\noindent The scheme \textit{Argument from inconsistency (tu quoque)} takes the following form:

\begin{scheme}
$P_1$: & Agent $X$ now claims or supports principle $P$ \\[2pt]
$P_2$: & Agent $X$ previously supported action~$A$ \\[2pt]
$P_3$: & Action $A$ and principle $P$ are not compatible \\

\midrule
$C$:   & Therefore, agent $X$ is hypocritical or inconsistent
\end{scheme}

\noindent Here, $P_1$ and $P_2$ are recoverable from the rhetorical 
question, the administration invokes health autonomy while having compelled 
experimental vaccination, but the bridging premise $P_3$, that compelling 
such vaccination violates that very right, is implicit. Without it, the charge 
of hypocrisy does not follow; the tweet relies on the reader to supply it. The conclusion here is also explicit. 

The following section formalizes how these schemes operate within 
our framework: once identified, their inferential license is reified as a 
propositional conditional and injected into the premise set, restoring 
classical entailment between the reconstructed content and the conclusion.

\subsection{Formal Definition of Enthymematic Arguments for Computational Analysis}
\label{sec:formal}

We adopt the notation of formal argumentation theory from \citet{besnardelements}, 
in which premises and conclusions are represented as formulae of a classical 
propositional language, and the inferential relation between them is the classical 
consequence relation $\vdash$. We first define the enthymeme in this setting, 
then show how argumentation schemes enter as reconstructed propositional content.

We represent an explicit argument as a pair $(P, c)$, where $P = \{p_1, \dots, p_n\}$ 
is a set of explicit premises and $c$ is a single conclusion. An argument is 
enthymematic whenever its explicit material alone is insufficient to establish 
a logically complete inference:
\begin{equation}
    P \nvdash c
\end{equation}

Enthymeme annotation consists in identifying a set of implicit elements 
$\Delta = \Delta_P \cup \Delta_R$, where $\Delta_P$ is the set of reconstructed 
implicit material premises, and $\Delta_R$ is the set of reconstructed 
propositional conditionals that encode the inferential licenses supplied by 
argumentation schemes \citep{walton2008}. While both $\Delta_P$ and $\Delta_R$ 
function as propositions to satisfy entailment, we maintain 
this structural distinction to separate domain-specific factual filling ($\Delta_P$) 
from argument schemes ($\Delta_R$). 

When $c$ is absent from the explicit material, $c'$ is additionally reconstructed 
as a single implicit conclusion $\delta_c$:
\[
c' = \begin{cases} c & \text{if } c \text{ is explicit} \\ 
\delta_c & \text{otherwise} \end{cases}
\]

\noindent Note that while Walton schemes are inherently defeasible and
non-monotonic, we follow a deductivist reconstruction approach,
translating scheme instances into implications to maintain a deductive
framework.\footnote{We are explicit about what the entailment requirement does
and does not accomplish. Taken in isolation it is nearly vacuous: for any
premises $P$ and conclusion $c$, adding the associated conditional
$(\bigwedge P) \rightarrow c$ restores classical entailment trivially --- the
standard objection to deductivist reconstruction recalled in
Section~\ref{sec:background-theory}. We therefore do not treat entailment as
the criterion that selects among candidate reconstructions. Its role is that of
a completeness check: it forces every proposition the inference depends on
to be made explicit, and makes that completeness mechanically checkable. The
substantive work is done elsewhere --- by the requirement that the reconstructed
conditional instantiate a recognized argumentation scheme whose slots are
filled by propositions available in the reconstructed argument
(Section~\ref{sec:walton-med}), and by the interpretive constraints of the
annotation guidelines (Section~\ref{sec:guidelines}).} In this framework,
entailment is only a mechanical completeness check: the substantive constraint
on what may be reconstructed comes from the scheme.

\subsubsection{Direct Entailment}

In the simplest case, the missing element is a propositional bridge that,
once supplied, makes the inference deductively valid without appeal to an
argumentation scheme ($\Delta_R = \emptyset$). In this configuration, a
reconstructed conditional functions as an ordinary material premise and is
therefore an element of $\Delta_P$; $\Delta_R$ is reserved for conditionals
that reify the inferential license of an argumentation scheme.

Missing premise. Consider the following example:
\begin{quote}
\textit{``Got sick from the very thing she got `vaccinated' for.. ergo, its not 
a classic vaccine."} \#298
\end{quote}
\noindent Here, $e_1$ is the only explicit premise and $e_2$ the explicit conclusion:
\begin{itemize}
    \item[$e_1$:] \textit{She got sick from the very thing she was vaccinated for}
    \item[$e_2$:] \textit{It is not a classic vaccine}
\end{itemize}
\noindent Since $\{e_1\} \nvdash e_2$, annotation reconstructs the implicit conditional that directly links them:
\[
    \delta_P : e_1 \rightarrow e_2
\]
\noindent (i.e., \textit{``If someone gets sick from the very thing they were vaccinated for, it is not a classic vaccine''}), yielding the reconstructed argument $(\{e_1, \delta_P\},\ e_2)$.

Missing conclusion. Consider the following example:
\begin{quote}
\textit{``It's revealing that lying about his asylum claim in 2015 was insufficient 
to remove Emad Al Swealmeen from the UK. He remained and detonated a bomb on 
Remembrance Sunday last year. It begs the question. If making false claims is 
not sufficient to remove people --- what is?"} \#3013
\end{quote}
\noindent The tweet conveys two premises but draws no explicit conclusion. Note
that $e_1$ is not asserted verbatim: it is conveyed by the closing rhetorical
question (``If making false claims is not sufficient to remove people --- what
is?''), which presents the sufficiency principle as beyond dispute and thereby
commits the speaker to it; following our guidelines
(Section~\ref{sec:guidelines-boundary}), content conveyed by rhetorical
questions is treated as explicit:
\begin{itemize}
    \item[$e_1$:] \textit{Making false claims should be sufficient grounds for removal}
    \item[$e_2$:] \textit{Swealmeen made false claims}
\end{itemize}
\noindent $e_1$ encodes the conditional $p \rightarrow q$ and $e_2$ encodes 
$p$, but no conclusion is explicitly drawn. Annotation reconstructs the 
implicit conclusion:
\[
    \delta_c : \textit{Swealmeen should have been removed}
\]
\noindent yielding the argument $(\{e_1, e_2\},\ \delta_c)$ where 
$\{e_1, e_2\} \vdash \delta_c$.

\subsubsection{Walton Scheme-Mediated Inference}
\label{sec:walton-med}

When the inferential step appeals to a stereotypical pattern of reasoning, 
an argumentation scheme supplies the inferential license. 
The scheme guides annotators toward the correct conditional to reconstruct.

Let $\mathcal{W}$ be a Walton argumentation scheme with proposition slots 
$X_1, \dots, X_n$. When the slots of $\mathcal{W}$ are filled by propositions 
in $P \cup \Delta_P \cup \{c'\}$, the scheme licenses the reconstruction of 
a propositional conditional:
\[
    \delta_R : \sigma(\mathcal{W}) \rightarrow c'
\]
\noindent where $\sigma(\mathcal{W})$ is the instantiation of the scheme's 
premises over the available propositions and $\rightarrow$ denotes
implication. $\delta_R$ is added to $\Delta_R$.

\paragraph{Missing premise.} Consider the following example:
\begin{quote}
\textit{``Or hope, Judy, was it hope? Tell us all about that `proven' science 
that was used to tell us the vaccine would prevent infection and transmission. 
Birx says it wasn't proven, it was just hope."} \#1047
\end{quote}
\begin{itemize}
    \item[$e_1$:] \textit{Birx says the vaccine preventing infection and 
    transmission was not proven}
    \item[$e_2$:] \textit{The vaccine was not proven}
\end{itemize}
\noindent Since $\{e_1\} \nvdash e_2$, the \textit{Argument from Position 
to Know} scheme is identified, whose abstract form is:
\[
    \underbrace{(S \text{ is in a position to know about } D)}_{\delta_P} 
    \wedge 
    \underbrace{(S \text{ asserts } p \text{ about } D)}_{e_1} 
    \rightarrow 
    \underbrace{p}_{e_2}
\]
\noindent Here, the conclusion $p$ maps to $e_2$. Annotation reconstructs the 
missing situational slot filler ($\delta_P$) as well as the structural conditional 
warrant ($\delta_R$):
\[
    \delta_P : \textit{Birx is in a position to know about vaccines}
\]
\[
    \delta_R : (\delta_P \wedge e_1) \rightarrow e_2
\]
\begin{sloppypar}
\noindent yielding the complete reconstructed argument
$(\{e_1, \delta_P, \delta_R\},\ e_2)$, where
$\{e_1, \delta_P, \delta_R\} \vdash e_2$.
\end{sloppypar}

Missing conclusion. Consider the following example:
\begin{quote}
\textit{``To stop the deaths of those taking dangerous, illegal journeys, we 
must stop the trade in people that causes them. Our New Plan for Immigration 
will save lives and target people smugglers."} \#2274
\end{quote}
\begin{itemize}
    \item[$e_1$:] \textit{Stopping people smugglers is desirable}
    \item[$e_2$:] \textit{The New Plan for Immigration will stop people 
    smugglers}
\end{itemize}
\noindent Since $\{e_1, e_2\} \nvdash c'$, the \textit{Argument from Goal} 
scheme is identified:
\[
  \begin{aligned}
    &\underbrace{(\text{Goal } G \text{ is desirable})}_{e_1}
     \wedge
     \underbrace{(\text{Action } A \text{ achieves } G)}_{e_2} \\
    &\qquad \rightarrow
     \underbrace{(\text{Action } A \text{ should be supported})}_{c'}
  \end{aligned}
\]
\noindent Annotation reconstructs the propositional conditional encoding the 
scheme's license:
\[
    \delta_R : (e_1 \wedge e_2) \rightarrow c'
\]
\noindent and the implicit conclusion filling the missing slot:
\[
    c' = \delta_c : \text{We should support the New Plan for Immigration}
\]
\noindent yielding the argument $(\{e_1, e_2, \delta_R\},\ \delta_c)$ where 
$\{e_1, e_2, \delta_R\} \vdash \delta_c$.

\subsubsection{Summary and Final Definition}

\begin{definition}[Enthymeme Reconstruction]
Given an explicit argument $(P, c)$, its formal enthymeme reconstruction is 
a tuple $(P \cup \Delta_P \cup \Delta_R, c')$, where $\Delta_P$ is a set of 
reconstructed implicit premises, $\Delta_R$ is a set of reconstructed structural 
conditionals, and $c'$ is the final explicit or implicit conclusion. 

\noindent The reconstruction is structurally complete if and only if it restores classical entailment:
\[
    (P \cup \Delta_P \cup \Delta_R) \vdash c'
\]
\noindent When the argument appeals to a Walton scheme $\mathcal{W}$, 
$\Delta_R = \{\delta_R\}$, where $\delta_R : \sigma(\mathcal{W}) \rightarrow c'$ 
is obtained by instantiating the scheme's premise structure $\sigma(\mathcal{W})$ 
such that all instantiated terms are elements of $P \cup \Delta_P$.
\end{definition}

\noindent The entailment condition thus certifies completeness, not adequacy:
a reconstruction satisfying it may still misread the text. Adequacy is
enforced by the scheme-instantiation clause and by the interpretive
requirements of the annotation guidelines (Section~\ref{sec:guidelines}).

\section{Data Collection}
\label{sec:data}

\subsection{Data Source}
\label{sec:data-source}

The data in the resource we present here is from a preexisting dataset of tweets introduced by \citet{Flaccavento2025} in their study of trope usage in social media. The authors collected approximately 50,000 English-language tweets via keyword-based queries targeting vaccines (\texttt{"vaccine"}, June 26--27, 2022) and immigration (\texttt{"migrant"}, \texttt{"migration"}, \texttt{"asylum"}, spanning a period of four years, from late 2019 to late 2023), without filtering for specific users. They subsequently cleaned the dataset by stripping links and removing personally identifiable information, without applying a profanity filter. From this collection, they manually labeled a subset of 3,300 tweets (2,074 vaccine-related, 63\%; 1,226 immigration-related, 37\%) across nine trope categories; this labeled subset constitutes their released dataset.

From this dataset, we randomly selected a subset of 1,482 tweets so as to obtain a balanced distribution of trope vs.\ no-trope labels.

\subsection{Data Statistics and Characteristics}
\label{sec:data-stats}

Our dataset thus comprises 1,482 tweets drawn from two topics: COVID-19 discourse 
(1,169 tweets, 78.9\%) and immigration (313 tweets, 21.1\%), a roughly 4:1 
imbalance reflecting the relative volume of originally collected data. Table~\ref{tab:corpus-stats} 
summarizes the main distributions.

Overall the tweets are short: the average length is 29 tokens (173 characters), with a near-identical median, indicating a symmetric distribution. Three quarters of tweets fall in the 11--40 token range, with the 31--40 token 
bracket being the most frequent (26.7\%). Immigration tweets are marginally longer on average (31 tokens, 195 characters) than COVID-19 tweets (28.5 tokens, 167 characters). Surface features are sparse: 7\% of tweets contain  hashtags, 5.1\% have mentions such as '@sajidjavid', '@CCastaner'; URLs are absent, having been stripped prior to annotation.

The vocabulary reflects the topic split. COVID-19 tweets cluster around 
\textit{vaccine}, \textit{covid}, \textit{experimental}, and \textit{mandates}, 
with distinctive words including \textit{pfizer}, \textit{mrna}, \textit{cdc}, 
and \textit{gates}. Immigration tweets center on \textit{immigration},
\textit{illegal}, and \textit{migrants}, with topic-unique words such as 
\textit{asylum}, \textit{brexit}, \textit{ukip}, and \textit{sajidjavid}, 
pointing to a predominantly UK political context. The higher type-token ratio 
for immigration tweets (0.235 vs.\ 0.145) reflects greater diversity, 
consistent with the smaller sub-dataset size.

\begin{table}[h]
\centering
\caption{Dataset statistics by topic}
\label{tab:corpus-stats}
\renewcommand{\arraystretch}{1.25}
\begin{tabular}{lrrr}
\toprule
& \textbf{Overall} & \textbf{COVID-19} & \textbf{Immigration} \\
\midrule
Tweets          & 1,482   & 1,169 (78.9\%) & 313 (21.1\%) \\
Avg.\ tokens    & 29.0    & 28.5           & 31.0         \\
Avg.\ characters & 173    & 167            & 195          \\
\midrule
Total tokens    & 42,597  & --             & --           \\
Type-token ratio & 0.138  & 0.145          & 0.235        \\
\bottomrule
\end{tabular}
\end{table}

\section{Annotation Framework and Guidelines Development}
\label{subsec:annotation-framework-development}

Developing the annotation guidelines for our enthymeme resource was carried out across four phases. The first phase, Phase~0, was an exploration phase in which two annotators (and a referee for disagreements) annotated 313 tweets using the labels \texttt{implicit\_premise}, \texttt{implicit\_conclusion}, and \texttt{none}, with a free-text box to record comments and their interpretation of the argument structure. A first version of the guidelines emerged from this phase.

Phase~1 was where most of the work in developing the guidelines took place, following its own iterative process that we detail in Section~\ref{subsubsec:iterative-process}. Phase~2 was about seeing how different and newly available annotators would engage with the guidelines in terms of understanding and clarity, and also to select the best possible annotators for Phase~3, the final phase in which the full resource was annotated.

\begin{figure}[ht]
\centering
\begin{tikzpicture}[
    font=\footnotesize,
    box/.style   ={draw=gray!65, line width=0.5pt, rounded corners=2pt,
                   fill=gray!6, minimum width=2.35cm, minimum height=0.52cm,
                   inner sep=2pt, font=\footnotesize\itshape},
    down/.style  ={-{Latex[length=3.5pt,width=3pt]}, gray!75, line width=0.6pt},
    hop/.style   ={-{Latex[length=3.5pt,width=3pt]}, gray!75, line width=0.6pt,
                   rounded corners=2.5pt},
  ]
  \def\colsep{2.72}
  \foreach \i/\ph in {0/0, 1/1, 2/2, 3/3}{
    \node[font=\small] at (\i*\colsep, 1.95) {Phase~\ph};
    \node[box] (a\i) at (\i*\colsep, 1.35) {Analysis};
    \node[box] (g\i) at (\i*\colsep, 0.60) {Guidelines};
    \node[box] (n\i) at (\i*\colsep, -0.15) {Annotate};
    \draw[down] (a\i) -- (g\i);
    \draw[down] (g\i) -- (n\i);
  }
  \foreach \i/\j in {0/1, 1/2, 2/3}{
    \draw[hop] (n\i.east) -- ++(0.25,0) |- (a\j.west);
  }
  \draw[gray!45, line width=0.4pt] (-1.28,-0.72) -- (3*\colsep+1.28,-0.72);
  \foreach \i/\n/\t in {%
      0/{313 examples}/{Immigration},
      1/{156 examples}/{Immigration},
      2/{75 + 75 examples}/{Immig.\ \& Covid},
      3/{313 + 1{,}169 examples}/{Immig.\ \& Covid}}{
    \node[anchor=north west, align=left, font=\footnotesize]
      at (\i*\colsep-1.15,-0.90) {$\cdot$~\n\\[1pt] $\cdot$~\t};
  }
\end{tikzpicture}
\caption{Overview of the four annotation phases}
\label{fig:annotation-phases}
\end{figure}

\subsection{Iterative Process (Phase 1) }
\label{subsubsec:iterative-process}

This phase had as its objective the development of the annotation guidelines. Master's students from the Linguistics Research Master volunteered for the guidelines development project. The 156 tweets for this phase were divided into four sets (36, 40, 40, and 40 tweets; see Figure~\ref{fig:annotation-phase1}), which the three annotators\footnote{Annotator identifiers used in this phase (Annotators 1--3) are local to the guideline-development phase and do not correspond to the identifiers Ann~1--5 used in the final annotation campaign (Section~\ref{sec:process}).} would work on independently before meeting after each set to discuss and analyze the tweets and their annotations.

\subsubsection{Multi-step Inductive Coding}
After some initial training, the annotators were instructed to approach this task using an inductive coding strategy. For each set, they were given the following instructions:

\begin{itemize}
    \item Analyze different cases for each label, that is different cases for \texttt{implicit\_conclusion}, \texttt{implicit\_premise}, and \texttt{none}.
    \item Come up with your own categories.
    \item Classify cases from easy to hard.
    \item For the next set: refine categories and repeat.
\end{itemize}

After the meeting following the first set, governing principles and essential criteria for identifying the kind of implicit content we were looking for began to emerge. The idea was not to come up with a multitude of precise annotation rules, but rather to identify the essential annotation criteria. Furthermore, after the first meeting, following deliberation over Set~1, it was decided that each annotator should supply a propositional breakdown of the tweet, placing its components into slots --- premises $p_1, p_2, p_3, \ldots$ and conclusion $c$ --- in order to more easily follow the reasoning and better understand the argument structure as interpreted by the annotator.

\subsubsection{Controversial Content Scope and Semantic Space}

One of the central and recurring points of discussion during Phase~1 concerned the question of what counted as controversial content --- more precisely, what kind of implicit claims could qualify as carrying controversial political content within the scope of our dataset. Some tweets reached beyond the Immigration and Covid topics to make more abstract claims about politics that were not 
anchored to either topic specifically, making it genuinely unclear whether they fell within the scope of what we considered controversial for our purposes. We settled on treating content as controversial when it has the potential to influence or manipulate opinion, the perception of a state of affairs, or behaviour with respect to COVID-19 and immigration.

It therefore became necessary to explicitly define the semantic space in which the annotation operated. This step corresponds to what \citet{Aroyo2015} identify as a prerequisite for any annotation campaign in which disagreement is to function as signal rather than noise: the semantic space must be bounded and its dimensionality constrained so that annotators share a common reference frame within which meaningful agreement and disagreement can occur. Without such 
delimitation, diverging annotations reflect not genuine interpretive differences but simply the absence of a shared scope.

To this end, we decomposed the Immigration and Covid topics into a set of subtopics, and for each subtopic provided examples of implicit controversial content that fell  within its scope. This gave annotators a concrete and bounded semantic space within which to situate their judgments. This process resulted in Table~\ref{tab:semantic-space} of the Appendix.

\begin{figure}[ht]
\centering
\includegraphics[width=\textwidth]{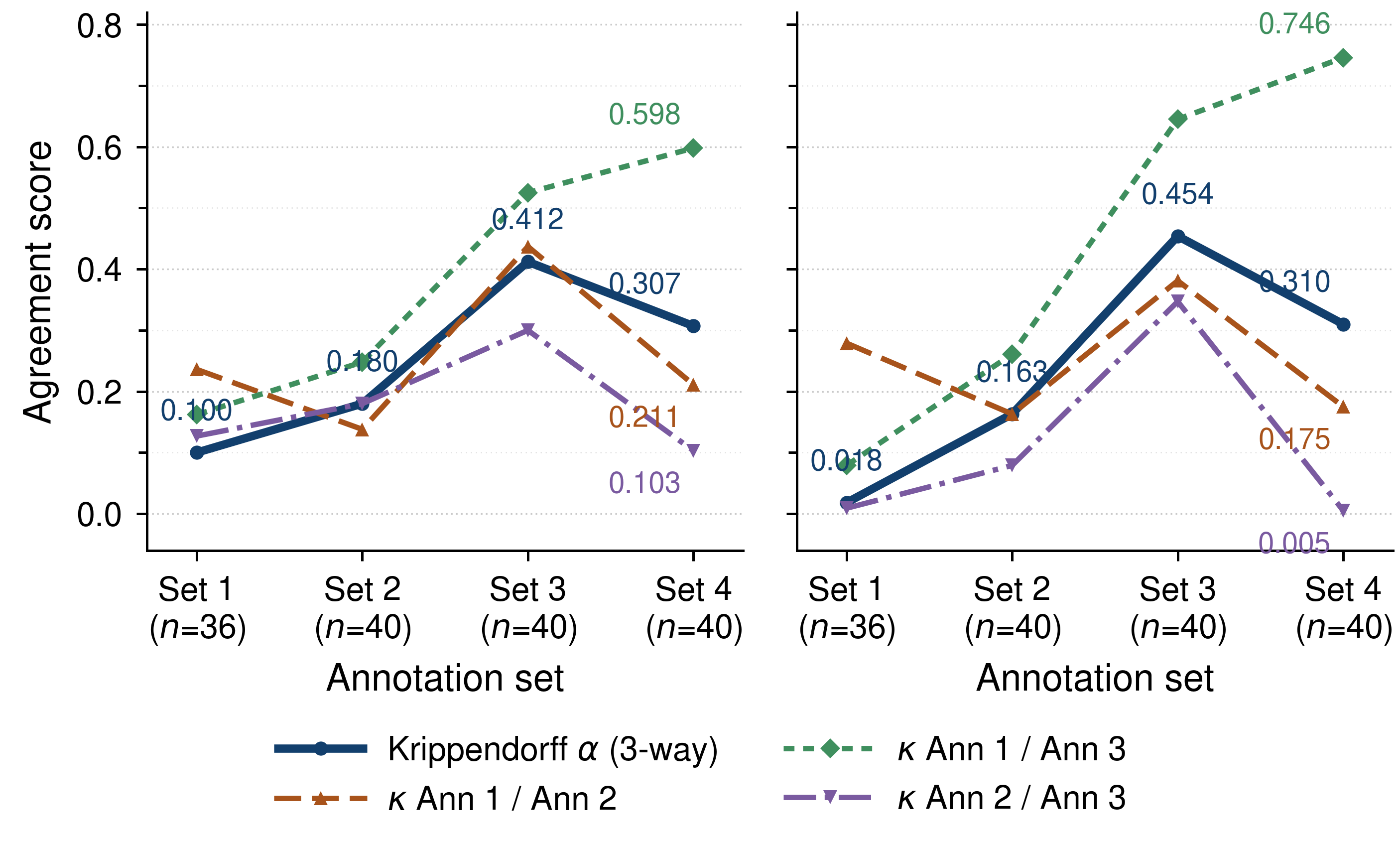}
\caption{Inter-annotator agreement (Krippendorff's $\alpha$ and pairwise
Cohen's $\kappa$) across annotation sets 1--4, for the original three-class
labels (left; \textit{premise} / \textit{conclusion} / \textit{none}) and for
the merged two-class labels (right; \textit{implicit} / \textit{none})}
\label{fig:annotation-phase1}
\end{figure}

\subsubsection{Inter-Annotator Agreement (IAA) over Time during Training}

Figure \ref{fig:annotation-phase1} shows how IAA evolved across the four annotation sets as the guidelines were iteratively refined. The overall trend is positive: on the merged two-class labels, three-way Krippendorff's $\alpha$ rises from 0.018 in Set 1 to 0.454 in Set 3, indicating moderate agreement. The three-class $\alpha$ follows the same trajectory at a lower level, from 0.100 to 0.412.

The first improvement occurs at Set 2, where the propositional breakdown constraint was introduced. Annotators were required to make their argument reconstruction explicit through structured free-text fields for premises ($p_1$, $p_2$, $p_3$) and the $conclusion$. This reduced interpretive latitude and grounded annotations in a shared argument structure. Work on argumentation schemes was also initiated at this stage: in the interpretation free-text field, annotators were required to name the scheme they used. If they could not identify it by name, they were asked to write it out in full premise-conclusion format, as described in Section~\ref{sec:schemes}. This pushed annotators to reflect more explicitly on the inferential structure underlying their reconstructions.

Agreement rises more sharply at Set 3, following the introduction of a controversial content scope constraint. Previously, guidelines only loosely indicated that the implicit content should concern immigration or COVID-19. Set 3 defined this semantic space more precisely, narrowing the range of acceptable reconstructions.

Set 4 shows a decline: despite $\kappa = 0.746$ for Ann 1 / Ann 3, three-way $\alpha$ drops to 0.310. Post-hoc review revealed that Annotator 1 showed inconsistent quality throughout all phases, not just Set 4. While they produced valid annotations at times, their quality would deteriorate in streaks — consecutive tweets annotated side by side where reconstructions were off-topic or non-compliant with the argument structure constraint. This pattern was particularly present towards the end of sets, suggesting attention or cognitive overload effects rather than a stable annotation strategy. This is a pattern of local systematic annotation errors rather than genuine disagreement, and it raises the question of where annotation error ends and legitimate label variation begins, which is central to our analysis of disagreement as signal. It also motivated our decision to exclude this annotator from the final annotation phase; they did not participate in the final campaign, whose identifier Ann~1 refers to a different participant (Section~\ref{sec:process}).

\subsection{Final Phase, Refining guidelines and Schemes}
\label{subsubsec:reference-set-and-schemes}

The final minor modifications to the guidelines were made during the first step of the final annotation phase (Phase 3). A reference set was annotated by all five selected annotators, followed by a consolidation discussion before annotating the rest of the resource. As detailed in Section~\ref{ann-design}, different splits were then assigned to different annotators given resource and timing constraints, with a minimum of three annotations per item guaranteed.

After the reference set, the last addition to the guidelines concerned argumentation schemes: the most frequently used and most recognizable schemes were documented and added as examples. The guidelines were not modified after this point.

\subsection{Annotation Schema and Label Set}

Each annotation instance comprises three layers. 
First, the annotator assigns a \textbf{categorical label}: \textit{implicit 
premise}, \textit{implicit conclusion}, or \textit{none} (no-enthymeme). 
Second, the annotator performs a \textbf{propositional decomposition}: the 
tweet is segmented into propositions $p1$, $p2$ (and optionally $p3$), and a 
conclusion, with one slot marked as implicit --- either one of the premises 
or the conclusion, consistent with the categorical label. Third, the annotator 
provides a \textbf{free-text reconstruction} of the implicit element using 
the scheme's variable notation; 
annotators need not identify a formal scheme by name.

\noindent An annotation for Tweet 2274 from Section~\ref{sec:walton-med} is shown
in Table~\ref{tab:instance-2274}.

\begin{table}[h]
\centering
\caption{Annotation instance for Tweet 2274}
\label{tab:instance-2274}
\begin{tabular}{lp{9cm}}
\toprule
\textbf{Field} & \textbf{Value} \\
\midrule
Label       & \textit{Implicit conclusion} \\
P1          & \textit{Stopping people smugglers is desirable} \\
P2          & \textit{The New Plan for Immigration will stop people smugglers} \\
Conclusion  & \textbf{We should support the New Plan for Immigration [implicit]} \\
Scheme      & \textit{Argument from Goal} \\
\bottomrule
\end{tabular}
\end{table}

\section{Annotation Complexity Analysis}
\label{sec:dimensions}

In this section, we analyze and measure the complexity of the annotation task for
annotating enthymemes as defined in the previous section. We follow the framework
proposed by \citet{fort-etal-2012-modeling}, which decomposes annotation complexity into
six dimensions. The first two, \textbf{Discrimination} and \textbf{Delimitation},
concern the work annotators must perform to localize \emph{what} is to be annotated. The
next three, \textbf{Expressiveness}, \textbf{Label Set} and \textbf{Ambiguity}, concern
\emph{how} annotators must proceed once a unit has been identified. The sixth and final
dimension addresses the role of \textbf{Context} in the annotation process.

Although annotators do not consciously address each of these six dimensions in this
particular order, the framework is nonetheless useful for managing annotation campaigns
\citep{fort2012ressources}: it provides an estimate of the cognitive load and time
required to complete an annotation task, and helps set realistic expectations given
finite material and human resources. Moreover, these dimensions offer a principled basis
for demonstrating that semantic interpretation tasks are inherently difficult---and thus
for explaining why moderate inter-annotator disagreement is typically expected in such
settings.

That last argument only carries weight if the scores are derived rather than asserted.
Each dimension is defined by an explicit formula whose inputs are quantities of our own
corpus, and it is the substitution of those quantities that turns an intuition about
difficulty into a measurement---which matters here because two of the six dimensions
reach their ceiling and one is reported as a range, results that would read as
stipulations if the inputs were not shown. We therefore report below the inputs driving
each score together with the reasoning connecting them to the result, and refer the
reader to \citet{fort-etal-2012-modeling} and \citet{fort2012ressources} for the formal
definition of each measure and the procedure for computing it
(summarized in Table~\ref{tab:complexity}).

\begin{table}[ht]
\centering
\caption{Complexity profile of enthymeme annotation across the six dimensions of
\citet{fort-etal-2012-modeling}, to which we refer for the formal definition of each
measure and the procedure for computing it}
\label{tab:complexity}
\small
\setlength{\tabcolsep}{5pt}
\begin{tabularx}{\textwidth}{@{}l X r@{}}
\toprule
\textbf{Dimension} & \textbf{Basis in our corpus} & \textbf{Score} \\
\midrule
Discrimination & 502 of 5,070 candidate units, across three segmentation levels,
                 require annotation & 0.90 \\
Delimitation   & 1,580 insertions and 1,078 substitutions over 1,580 annotation
                 units; ceiling reached & 1.0 \\
Expressiveness & arity ${>}2$ relational language for structure, higher-order for
                 reconstruction & 0.75--1.0 \\
Label set      & $\nu \leq 5$ against $\tau = 50$ for surface labels; unbounded for
                 reconstruction & 0.10 / 1.0 \\
Ambiguity      & weighted uncertainty markers over 2,808 rated annotations & 0.45 \\
Context        & tweet-level text span, unidentified external sources & 1.0 \\
\bottomrule
\end{tabularx}
\end{table}

\subsection{Analysis by Dimension}
\label{sec:dimensions-analysis}

\paragraph{Discrimination.}
Discrimination captures the cost of identifying what to annotate prior to the act of
labeling itself, and can be as demanding as the labeling: in semantic role labeling or
discourse annotation the relevant unit of segmentation may fall below the word level or
above the sentence level \citep{erk-etal-2003, widlocher-mathet-2009}, whereas it is a
non-issue in part-of-speech tagging, where every word must receive a label
\citep{fort2012ressources}. Although the tweet as a whole constitutes our unit of
analysis, annotators must perform three logically ordered sub-tasks: determining whether
the tweet contains an argument worth investigating; reconstructing the relevant
propositions, including any implicit component; and assigning each proposition its
enthymematic role. None of these units are pre-segmented in the text. Segmenting the
dataset at the three corresponding levels yields 1,482 tweets, some 1,580 candidate
propositional units and 2,008 argument component slots, of which only the 502
argumentative tweets identified at the first level ultimately require annotation. The
resulting score of $\mathbf{0.90}$ indicates that discrimination is a major source of
complexity: only a small fraction of all candidate units across all segmentation levels
requires annotation, yet identifying them demands sustained interpretive effort at each
level.

\paragraph{Delimitation.}
Discrimination alone is not sufficient for localization in textual data: the units it
identifies still require their boundaries to be established through delimitation, which
accounts for all separation and unification operations applied to the basic annotation
unit. Across the 502 tweets containing enthymemes, every tweet is first fully decomposed
into its propositional units, giving some 1,580 insertions, with the implicit
proposition counting as one structurally guaranteed unit within that figure; every
explicit component---all components except the one implicit proposition per tweet, 1,078
units in total---then requires the annotator to trim and delimit its textual boundaries,
selecting exactly which words belong to that logical proposition; and no merging of
units was required across the dataset. Normalized by the 1,580 annotation units
produced, the measure hits its maximum value of $\mathbf{1.0}$. This was to be expected
given the addition of implicit content that extends beyond the basic units constituted
by the tweets: the cap at 1 was introduced precisely to anticipate such cases, so a
score of 1.0 means that the task is at maximum complexity along this dimension.

\paragraph{Expressiveness.}
Expressiveness concerns the labels that can be assigned to the discriminated and
delimited units, or to the relations between them, graded on a four-point scale running
from type languages (0.25) through relational languages of arity 2 (0.50) and of arity
${>}2$ (0.75) to higher-order languages (1). The final subtask---writing a natural
language proposition that captures the implicit component---qualifies as a higher-order
language and scores 1. Assigning surface labels (\textit{implicit premise},
\textit{implicit conclusion}, \textit{none}) and determining the argument structure
score 0.75: they involve relational annotation of arity ${>}2$, since the annotator
constructs a typed relation between multiple typed components ($p_1$, $p_2$, optionally
$p_3$, and a conclusion) and assigns an argumentation scheme to that multi-element
structure, which is precisely the ``who did what to whom, when, and how'' class of
annotation associated with arity ${>}2$. Even restricting attention to surface labels and
argument structure alone, enthymeme annotation therefore scores high in expressiveness,
between $\mathbf{0.75}$ and $\mathbf{1.0}$.

\paragraph{Label set.}
Label set complexity is about the combinations of possible choices given a finite label
set, relating the global degree of freedom $\nu$ that the annotator has for a simple or
complex label to a threshold $\tau$ at which a label set is considered to exceed what is
practically interpretable. $\tau$ is set to 50, a number empirically derived from
\citeauthor{fort2012ressources}'s (\citeyear{fort2012ressources}) experiments on various
linguistic annotation campaigns. As with delimitation, the reconstruction task
immediately reaches the maximum score and warrants separate treatment. In assigning one
of the three labels, the annotator goes through two sequential decisions whose order does
not matter: placing the implicit argument component---an implicit conclusion, or which of
\textit{p1}, \textit{p2}, \textit{p3} it occupies---which yields three degrees of
freedom, and ordering the explicit components across the up-to-three positions that
remain, which yields two more. The global degree of freedom is therefore bounded by $\nu
\leq 5$, well under the threshold, and the dimension scores $\mathbf{0.10}$. For the
reconstruction, the annotator does not select from a predetermined label set but writes a
natural language proposition from scratch; the space of possible choices exceeds $\tau$
at the very first word, so this sub-task scores $\mathbf{1.0}$.

\paragraph{Ambiguity.}
Our task is by nature ambiguous in the sense that multiple interpretations are possible
for a given tweet, yet the annotator is expected to select the most plausible one. This
makes ambiguity difficult to measure: since the annotator resolves it before committing
to an annotation, only one final interpretation is visible in the end.
\citet{fort2012ressources} proposes measuring what she calls residual ambiguity instead,
by monitoring the traces annotators leave behind in their disambiguation work. In our
campaign those traces take the form of what she refers to as uncertainty markers: our
annotators were asked to rate the difficulty of each of their annotations on a scale from
\textit{easy} to \textit{hard}, and residual ambiguity is the proportion of annotations
weighted by that rating, with \textit{easy} contributing 0, \textit{medium} 0.5 and
\textit{hard} 1. Over the 2,808 annotations carrying a rating---the subset of the roughly
4,800 original annotations collected in the campaign (Section~\ref{ann-design}) for which
it was recorded, as the rating was not used consistently by all annotators---we obtain
$\mathbf{0.45}$. As with the reconstruction, this dimension is difficult to measure in
practice: the label space is infinite by design and the available metric does not operate
on semantic similarity. A metric designed for open-vocabulary tasks would be needed here.

\paragraph{Context.}
The context dimension plays a role in each of the preceding complexity factors but is
represented separately here. Still following \citet{fort2012ressources}, the two criteria
to consider are the size of the source text needed to annotate a unit and the
accessibility of the knowledge resources needed to reach a decision; for consistency with
the preceding criteria, these two qualitative scales are translated into a single
discrete scale ranging from 0 to 1, represented in Figure~\ref{fig:context-weight} of
the Appendix. Concerning the text
span, the relevant unit at every stage of the annotation---assigning the surface label,
determining the argument structure, or writing the proposition capturing the missing
component---is the tweet itself: the enthymeme, when present, is contained within it, and
annotators have access neither to the broader thread nor to any tweet it may be
responding to. Where this introduces ambiguity, as with an explicit reply or a pronoun
that cannot be resolved without that context, the tweet is automatically annotated as
containing no enthymeme. Concerning the knowledge resources required, the task falls into
the category of unidentified external sources: tweet authors draw on a wide diversity of
linguistic registers, wordplay, cultural references and reactions to current events that
cannot be anticipated or grouped into a predefined set of documents, and annotators will
in many cases need to consult news articles, blog posts or Wikipedia entries that cannot
be specified in advance. Plotting the intersection of these two criteria on
Figure~\ref{fig:context-weight} of the Appendix---tweet-level text span on the vertical
axis, unidentified external sources on the horizontal axis---places our task at a context
weight of $\mathbf{1.0}$.

\subsection{Synthesis}
\label{sec:dimensions-synthesis}

\begin{figure}[ht]
\centering
\begin{tikzpicture}[x=1cm,y=1cm,
    ring/.style     ={gray!55, line width=0.3pt},
    recon/.style    ={black, line width=1.1pt},
    reconun/.style  ={black, line width=1.1pt, line cap=round,
                      dash pattern=on 0pt off 3.2pt},
    surf/.style     ={g75, line width=1.0pt,
                      dash pattern=on 3.6pt off 2.2pt}]
  \def\R{2.55}

  \fill[gray!7]
    (90:0.90*\R) -- (30:1.0*\R) -- (-30:1.0*\R) -- (-90:1.0*\R)
    -- (-150:1.0*\R) -- (150:1.0*\R) -- cycle;
  \fill[gray!20]
    (90:0.90*\R) -- (30:1.0*\R) -- (-30:0.45*\R) -- (-90:0.10*\R)
    -- (-150:0.75*\R) -- (150:1.0*\R) -- cycle;

  \foreach \r in {0.2,0.4,0.6,0.8,1.0}{%
    \draw[ring] (90:\r*\R) -- (30:\r*\R) -- (-30:\r*\R) -- (-90:\r*\R)
                -- (-150:\r*\R) -- (150:\r*\R) -- cycle;}
  \foreach \a in {90,30,-30,-90,-150,150}{\draw[ring] (0,0) -- (\a:\R);}

  \draw[recon]   (150:1.0*\R) -- (90:0.90*\R) -- (30:1.0*\R);
  \draw[reconun] (30:1.0*\R) -- (-30:1.0*\R) -- (-90:1.0*\R);
  \draw[recon]   (-90:1.0*\R) -- (-150:1.0*\R) -- (150:1.0*\R);
  \draw[surf]
    (90:0.90*\R) -- (30:1.0*\R) -- (-30:0.45*\R) -- (-90:0.10*\R)
    -- (-150:0.75*\R) -- (150:1.0*\R) -- cycle;

  \foreach \a/\r in {90/0.90, 30/1.0, -90/1.0, -150/1.0, 150/1.0}{%
    \fill[black] (\a:\r*\R) circle (2.1pt);}
  \draw[black, line width=0.9pt, fill=white] (-30:1.0*\R) circle (2.4pt);
  \foreach \a/\r in {-30/0.45, -90/0.10, -150/0.75}{%
    \draw[g75, line width=0.85pt, fill=white]
      ($(\a:\r*\R)+(-2.3pt,-2.3pt)$) rectangle ($(\a:\r*\R)+(2.3pt,2.3pt)$);}

  \foreach \r in {0.2,0.4,0.6,0.8,1.0}{%
    \node[font=\tiny, text=gray!85, inner sep=0pt, anchor=east, xshift=-3pt]
      at (90:\r*\R) {\r};}

  \node[font=\small\bfseries, above=6pt]  at (90:\R)   {Discrimination};
  \node[font=\small\bfseries, right=7pt]  at (30:\R)   {Context};
  \node[font=\small\bfseries, right=7pt]  at (-30:\R)  {Ambiguity};
  \node[font=\small\bfseries, below=6pt]  at (-90:\R)  {Label set};
  \node[font=\small\bfseries, left=7pt]   at (-150:\R) {Expressiveness};
  \node[font=\small\bfseries, left=7pt]   at (150:\R)  {Delimitation};

  \node[anchor=north, font=\footnotesize] at (0,-3.25) {%
    \begin{tabular}{@{}c@{\hspace{0.7em}}l@{}}
      \dimkeyrecon  & Implicit proposition reconstruction \\[1pt]
      \dimkeysurf   & Argument structure and surface labels \\[1pt]
      \dimkeyunmeas & Not directly measurable \\
    \end{tabular}};
\end{tikzpicture}
\caption{Annotation complexity profile for enthymeme annotation across the six
dimensions of \citet{fort2012ressources}. The two series are given in the legend;
the dotted segment on the Ambiguity axis marks a value that cannot be measured
directly with existing metrics}
\label{fig:dimensions}
\end{figure}

Figure~\ref{fig:dimensions} shows that enthymeme annotation obtains high scores across
all dimensions, with two of them at their ceiling: delimitation and context both reach
1.0, the former as a structural consequence of the need to write an implicit
natural-language proposition that is not present in the original explicit text, the
latter because of the unanticipated external knowledge required to interpret political
tweets. The profile also makes the split within the task visible. Label set is the one
dimension that stays low, at 0.10, for surface-level labeling---yet it too jumps to 1.0
the moment an annotator must formulate the implicit argument component, which is why the
reconstruction is plotted separately throughout.

Overall, enthymeme annotation is highly demanding at almost every stage of the process:
identifying what to annotate, determining its boundaries, and producing the annotation
itself all require substantial cognitive effort. This is structural, as enthymeme
annotation is fundamentally an interpretative task that escapes purely mechanistic
procedures. The moderate inter-annotator agreement generally observed in such tasks
(Table~\ref{tab:resources}) is a direct consequence of the complexity demonstrated
and documented here.

\section{Annotation Guidelines}
\label{sec:guidelines}
The following subsections provide a summary of the annotation guidelines; readers are referred to the full guidelines\footnote{The full annotation guidelines are distributed together with the dataset; see the Data Availability statement.} for complete definitions and worked examples.

\subsection{Core Annotation Procedure}
\label{sec:guidelines-procedure}

Annotation proceeds in three sequential steps. First, the annotator assigns 
a categorical label: \textit{implicit\_premise}, \textit{implicit\_conclusion}, 
or \textit{none}. Second, the annotator segments the tweet into explicit 
propositions (P1, P2, and optionally P3) and a conclusion, marking which 
slot is implicit. Third, the annotator reconstructs the implicit element in 
free text, writing the argument scheme as a conditional over the proposition 
variables (e.g., \textit{if P1 and P2, then C}); no formal scheme name is 
required.

For an annotation to be valid, four criteria must be met:
\begin{enumerate}
    \item The missing component must express a stance on one of the topics 
    defined in Section~\ref{sec:guidelines-semantic};
    \item At least two premises must share a term;
    \item The reconstructed argument must instantiate a recognizable scheme;
    \item At least two argument segments must be explicit in the tweet.
\end{enumerate}
\noindent Only one implicit proposition may be reconstructed; additional 
implicit propositions are permitted only if they constitute commonsense premises which are necessary for the argument. 

\subsection{Semantic Space and Topic Definition}
\label{sec:guidelines-semantic}

We restrict annotation to tweets that (1) address one of two topics --- 
\textit{Immigration} and \textit{COVID-19} --- and (2) express a 
non-neutral stance. The missing component must itself carry a stance; 
neutral reconstructions are labeled \textit{none} regardless of how well 
they complete the argument structure. Each topic encompasses a range of 
sub-topics and associated stances, detailed in the full guidelines.

\subsection{Positive Examples: When to Annotate}
\label{sec:guidelines-positive}

The missing component may be a deliberate rhetorical choice, the 
unstated point being the argument's ``climax", or an assumption so
common within a discourse community that it goes without saying. In 
both cases, the criterion is persuasive potential: the implicit element 
should have a potential for influencing convictions or manipulating the audience's 
perception of a state of affairs. 
\citep{lombardivallauri2020implicit,macagno2022argumentation}.

Consider the example borrowed from 
\citeauthor{burke_vlah_2023} (\citeyear{burke_vlah_2023}):
\textit{``I don't want to get the COVID vaccine because 
one day I want to have babies.''}
A valid annotation is shown in Table \ref{tab:instance}.

\begin{table}[h]
\centering
\caption{Annotation instance for the vaccine example}
\label{tab:instance}
\begin{tabular}{lp{8.5cm}}
\toprule
\textbf{Field} & \textbf{Value} \\
\midrule
Label       & \textit{implicit\_premise} \\
P1          & \textbf{Vaccines cause infertility [implicit]} \\
P2          & \textit{I want to have babies} \\
Conclusion  & \textit{I don't want to get the COVID vaccine} \\
Scheme      & \textit{Argument from Sign} \\
\bottomrule
\end{tabular}
\end{table}

\subsection{Boundary Cases: When NOT to Annotate}
\label{sec:guidelines-boundary}

Several surface configurations resemble enthymemes but belong to distinct rhetorical phenomena that do not involve argumentative inference. Full discussion and examples for each case are provided in the detailed annotation guidelines; we summarize them here.

\textbf{Very likely logical consequences} are propositions entailed by the explicit text that 
add no argumentative
content, as are conditionals that merely restate the inferential step without
expressing a contentful generalization. \textbf{Circular and tautological arguments} involve premises and 
conclusions that express roughly the same proposition under different 
wording. \textbf{Rhetorical questions} convey an implied stance but do not 
constitute an argument scheme and are treated as explicit. 
\textbf{Coreference and anaphora resolution} recovers content already 
present in the text and therefore does not count as a missing component. 
\textbf{Neutral statement reconstruction} occurs when the implicit element 
completes the argument structure but does not express a stance on the 
defined topics; such cases are labeled \textit{none}. Finally,
\textbf{insufficient explicit segments} arise when fewer than two argument 
segments are explicit, requiring more than one implicit proposition to form
a complete argument, which exceeds our reconstruction limit.

\section{Annotation Process}
\label{sec:process}

Five annotators and two validators took part in the enthymeme annotation campaign. The annotators produced original annotations from scratch. The validators reviewed those annotations. The decision to introduce a validation tier emerged from practical constraints: given limitations in time and human resources, not all annotators could complete their assigned batches in full. To ensure a consistent level of annotation coverage, we adopted a design in which every datapoint received at least three independent original annotations and two subsequent validations. The details of this campaign can be seen in Section~\ref{ann-design}.

The present section describes the choices made to produce a robust enthymeme resource under constraints of time, human effort, and budget. Following the training phase, annotators were estimated to complete approximately 7.5 annotations per hour. For a target of 1,482 datapoints, full coverage by a single annotator would require roughly 198 hours of annotation work. Three of the five primary annotators received financial compensation, representing an initially estimated workload of 594 compensated hours.

The campaign ran over four months, during which annotators proceeded at varying speeds. Several annotators held concurrent employment or coursework obligations, which prevented some from completing their full 198-hour allocation in time. These circumstances necessitated adjustments to the original design, the outcome of which is reflected in Table~\ref{splits}. The following sub-sections discuss how these constraints were addressed and reflects on the implications of the resulting design choices for the quality and coverage of the resource.

\subsection{Annotator Profiles and Training}

Five annotators and two validators took part in the annotation campaign. The annotators were one lead expert annotator and four further annotators, three of whom were recruited from the university and financially compensated for their work.  Eligibility was established through a dedicated qualification phase, during which candidates completed a screening task to assess their suitability for the annotation task. Full professional proficiency in English was a strict prerequisite for participation, as all 
data to be annotated as well as the annotation guidelines were in English. The annotators' profile is detailed in Table \ref{tab:annotators}.

\begin{table}[h]
\centering
\caption{Summary profile of the annotators and validators}
\label{tab:annotators}
\small
\begin{tabular}{ll}
\toprule
\textbf{Dimension} & \textbf{Profile} \\
\midrule
Number of annotators & 7 (5 annotators + 2 validators)\\
Gender distribution  & 4 male, 3 female \\
Age range            & 24--60+ \\
Nationalities        & Dutch, French, Iranian, Chinese \\
Academic backgrounds & Computational linguistics, linguistics, \\
                     & data science, neuroscience, media studies \\
Seniority levels     & PhD / professorial (2), Master's level (4), \\
                     & industry professional (1) \\
English proficiency  & Full professional proficiency (mandatory) \\
Recruitment          & University, with financial compensation, Volunteer\\
Selection            & Qualification task (Phase 2) \\
\bottomrule
\end{tabular}
\end{table}

Prior to the main annotation campaign, all annotators went through a 
 training procedure. First, they attended a two-hour introductory session on argumentation theory, providing the conceptual 
grounding necessary for the task. Second, the annotation guidelines were 
reviewed collectively under the supervision of the expert annotator, who walked participants through the guidelines and addressed questions of interpretation, ensuring a shared understanding of the annotation categories and decision criteria.

Following this preparatory phase, annotators were assigned a training batch, which they annotated independently. The resulting annotations were then reviewed and given feedback by the expert annotator, with a particular focus on identifying and resolving systematic errors. This calibration step was designed to align annotator judgments before the main campaign and to prevent systematic misunderstandings of the guidelines early on.

\subsection{Annotation Campaign Design}
\label{ann-design}

\begin{table}[ht]
\centering
\caption[Tweet coverage by annotator and dataset split]{Tweet coverage by
annotator and dataset split. \spKey{splitprim}~primary annotation,
\spKey{splitcross}~cross-validation (by annotator),
\spKey{splitext}~external validation}
\label{splits}
\small
\setlength{\tabcolsep}{4.5pt}
\renewcommand{\arraystretch}{1.3}
\begin{tabular}{@{}l ccccc @{\hspace{1.4em}} cc@{}}
\toprule
 & \multicolumn{5}{c}{\textbf{Annotators}} & \multicolumn{2}{c}{\textbf{Validators}} \\
\cmidrule(lr){2-6}\cmidrule(l){7-8}
\textit{Split} & \textit{Ann 1} & \textit{Ann 2} & \textit{Ann 3} & \textit{Ann 4}
 & \textit{Ann 5} & \textit{Val 1} & \textit{Val 2} \\
\midrule
Split A $\cdot$ Set 1 & \spPrim{186} & \spPrim{186} & \spPrim{186} & \spCross{186} & \spNone & \spNone       & \spExt{186} \\
Split A $\cdot$ Set 2 & \spPrim{185} & \spPrim{185} & \spPrim{185} & \spCross{185} & \spNone & \spExt{185}   & \spNone \\
\addlinespace[2pt]
Split B $\cdot$ Set 1 & \spNone & \spCross{185} & \spPrim{185} & \spPrim{185} & \spPrim{185} & \spExt{185} & \spNone \\
Split B $\cdot$ Set 2 & \spNone & \spCross{185} & \spPrim{185} & \spPrim{185} & \spPrim{185} & \spExt{185} & \spNone \\
\addlinespace[2pt]
Final $\cdot$ Set 1   & \spNone & \spPrim{185} & \spPrim{185} & \spPrim{185} & \spNone & \spExt{185} & \spExt{185} \\
Final $\cdot$ Set 2   & \spNone & \spPrim{185} & \spPrim{185} & \spPrim{185} & \spNone & \spExt{185} & \spExt{185} \\
Final $\cdot$ Set 3   & \spNone & \spPrim{185} & \spPrim{185} & \spPrim{185} & \spNone & \spExt{185} & \spExt{185} \\
\midrule
\textit{\textbf{Reference}} & \spPrim{186} & \spPrim{186} & \spPrim{186} & \spPrim{186} & \spPrim{186} & \spNone & \spNone \\
\bottomrule
\end{tabular}
\end{table}

\subsubsection{Annotation Splits and Validation}
\label{annsplits}

The initial plan was to have all five annotators complete all data points, as Annotator (Ann) 3 did. Rapidly, however, given time constraints, varying annotator speeds, and overall progress relative to the deadline, compromises had to be made. Indeed, all annotators found that they could not fulfill their commitment to annotating the full dataset. Although the allocated hours would theoretically have fit their schedules (which already included professional obligations for some and coursework for others) most found that annotating for more than two hours a day was excruciating, given the high cognitive load the task demanded. Annotators consistently reported a significant drop in both their annotation capacity and quality after the two-hour mark.

As can be seen in Table~\ref{splits}, Ann 2 and Ann 4, who were the most efficient, were thus assigned the role of validators only for the splits they would not have time to annotate from scratch. Ann 1 and Ann 5, for their part, had to step down and limit themselves to the reference set and their respective splits.

Nevertheless, even with these compromises, the full dataset contains at least three original annotations per data point (five on the 186-tweet reference set), enabling us to study disagreement thoroughly, not only at the label level, but also in terms of what annotators see as the reasoning behind that label, revealing cases where annotators select the same label but for different reasons.

The validation phase allowed us to collect two additional validation annotations per data point, with the objective of either confirming or downweighting interpretations that might be erroneous. It was indeed reported that assessing compliance with the guidelines is considerably easier from an external standpoint than when one is actively reconstructing argument structures and implicit meanings in series. Validators were asked to either write their own interpretation drawing on the annotations already provided by the three original annotators, or select the annotation they deemed most sound and most consistent with the guidelines.

Validators agree with the original annotators more often than the original
annotators agree among themselves, since they choose among reconstructions
that already exist rather than producing one from scratch. The validation
layer therefore moves a proportion of contested items toward a majority
reading. We do not treat this as evidence that the majority reading is
correct: validators saw the original annotations before deciding, and the
agreement figures in Section~\ref{sec:statistics} are computed over original
annotations only.

\subsection{Annotation Tool and Data Presentation}

Each annotator received their data points in the form of a CSV file containing 185 or 186 tweets from the splits detailed above. To facilitate the annotation process, a dedicated annotation tool with a graphical user interface was built, allowing annotators to work directly within the tool rather than in MS Excel or some other spreadsheet application, which would have made the task considerably harder. All they had to do was upload their CSV to the application, which was made available as a website developed in native HTML and JavaScript, complete their annotations, and then download the finished result.

The interface presents the datapoints of a batch as a list from which the annotator may select tweets and annotate them in any order, and displays alongside the selected tweet the fields required at each step: difficulty rating buttons, five free-text boxes in which the argument structure is reconstructed by writing the propositions separately into their designated premise or conclusion slots, a scheme box in which the annotator names the identified argument scheme or writes out its abstract form, and finally the categorical label. Implicit propositions must be flagged manually by appending the string `(implicit)' after the proposition text. A screenshot of the interface and a walkthrough of the annotation workflow are included in the annotation guidelines, which are released together with the dataset and the tool itself (see the Data Availability statement).

\section{Final Dataset Statistics}
\label{sec:statistics}

\subsection{Label Distribution}

\begin{table}[ht]
\centering
\caption{Label distribution and annotator agreement strength per label
($N = 1{,}482$). Each bar shows the share of votes supporting the
majority label, from unanimous (5--0) to tiebreak (2--2--1). Vote counts
pool the three original annotations and the two validation votes per item
(the 186-tweet reference set carries five original annotations; see
Section~\ref{ann-design}). The tiebreak
share for \textit{none} is 0.0\%.
\textit{Conclusion} is the most contested label (11.1\% unanimous,
17.5\% tiebreak); \textit{none} the most stable (69.0\% unanimous)}
\label{tab:label-dist-agreement}
\small
\setlength{\tabcolsep}{6pt}
\renewcommand{\arraystretch}{1.55}
\begin{tabular}{l r r l}
\toprule
\textbf{Label} & \textbf{$N$} & \textbf{\%} & \textbf{Annotator agreement} \\
\midrule
\textit{none}       & 980     & 66.1  & \agreebar{69.0}{24.4}{6.6}{0.0} \\
\textit{premise}    & 439     & 29.6  & \agreebar{34.9}{41.0}{18.5}{5.7} \\
\textit{conclusion} &  63     &  4.3  & \agreebar{11.1}{36.5}{34.9}{17.5} \\
\midrule
Total               & 1{,}482 & 100.0 & \\
\bottomrule
\end{tabular}

\smallskip\noindent
\agreekeyplain{g75}\; Unanimous (5--0)\quad
\agreekey{g50}{north east lines}{g75}\; Strong (4--1)\quad
\agreekey{g28}{north west lines}{g50}\; Split (3--2)\quad
\agreekey{g12}{crosshatch}{g28}\; Tiebreak (2--2--1)

\end{table}

Table~\ref{tab:label-dist-agreement} shows that the dataset is moderately imbalanced, with \textit{none} accounting for two thirds of all items (66.1\%). The binary split between enthymeme and non-enthymeme cases follows a 1:2 ratio. Within the enthymeme subset, implicit premises are much more frequent than implicit conclusions (87.5\% vs.\ 12.5\%). This is consistent with what we know about how enthymemes work in practice: speakers more commonly leave a premise unsaid than a conclusion, since withholding a conclusion is a more marked rhetorical choice.

The vote counts underlying Table~\ref{tab:label-dist-agreement} pool the three original annotations with the two validation votes each item received (only the 186-tweet reference set carries five original annotations; see Section~\ref{ann-design}): a validator who endorsed an existing annotation contributes one vote to that annotation's label, and a validator who wrote their own interpretation contributes a vote to its label. Since validators saw the original annotations before deciding, validation votes are not fully independent; we include them here to give every item a uniform five-vote profile, while the inter-annotator agreement figures reported below are computed over original annotations only.

\subsection{Inter-Annotator Agreement}

\begin{table}[ht]
\centering
\caption{Inter-annotator agreement per batch. $\alpha$ is Krippendorff's alpha;
PA is pairwise percent agreement. The pooled row aggregates all non-reference
batches; the reference batch involves all five annotators}
\label{tab:iaa}
\small
\setlength{\tabcolsep}{7pt}
\renewcommand{\arraystretch}{1.2}
\begin{tabular}{l r l c c c}
\toprule
\textbf{Batch} & \textbf{$n$} & \textbf{Annotators} & $\bm{\alpha}$ \textbf{(3-class)} & $\bm{\alpha}$ \textbf{(binary)} & \textbf{PA} \\
\midrule
Split A $\cdot$ Set 2   & 185 & Ann 1, 2, 3       & 0.753 & 0.759 & 91.0\% \\
Split A $\cdot$ Set 1   & 186 & Ann 1, 2, 3       & 0.693 & 0.712 & 85.7\% \\
Final $\cdot$ Set 1     & 185 & Ann 2, 3, 4       & 0.544 & 0.571 & 80.5\% \\
Final $\cdot$ Set 3     & 185 & Ann 2, 3, 4       & 0.381 & 0.395 & 72.8\% \\
Final $\cdot$ Set 2     & 185 & Ann 2, 3, 4       & 0.324 & 0.335 & 74.8\% \\
Split B $\cdot$ Set 1   & 185 & Ann 3, 4, 5       & 0.280 & 0.388 & 56.8\% \\
Split B $\cdot$ Set 2   & 185 & Ann 3, 4, 5       & 0.218 & 0.324 & 51.9\% \\
\midrule
Pooled (non-ref.)       & 1{,}296 & 3-way          & \textbf{0.453} & 0.516 & 73.4\% \\
Reference (5-way)       & 186     & Ann 1, 2, 3, 4, 5 & 0.367 & 0.459 & 63.7\% \\
\bottomrule
\end{tabular}
\end{table}

From Table~\ref{tab:iaa}, we first observe that overall agreement measured with Krippendorff's
$\alpha$ \citep{krippendorff2011computing} is moderate overall, reaching substantial levels in the strongest batches, which is consistent with expectations for tasks requiring semantic interpretation. The binary collapse consistently yields higher agreement than the three-class task across all batches, confirming that \textit{conclusion} is the hardest label to agree on, as already suggested by the vote strength distribution in Table~\ref{tab:label-dist-agreement}.

We further note that agreement varies considerably across batches, with Krippendorff's
$\alpha$ ranging from 0.218 to 0.753 on the same task under the same guidelines. As we discuss below, this variation is largely explained by the composition of the team of annotators rather than content in different batches. Finally, the gap between binary and three-class
$\alpha$ widens in the lower-agreement batches: in batch A2 the difference is merely 0.006, while in batch B1 it reaches 0.108. This suggests that in more difficult annotation contexts, the premise/conclusion distinction breaks down first, annotators agree that an enthymeme is present, but diverge on which element is implicit.

\begin{figure}[ht]
\centering
\includegraphics[width=1.0\textwidth]{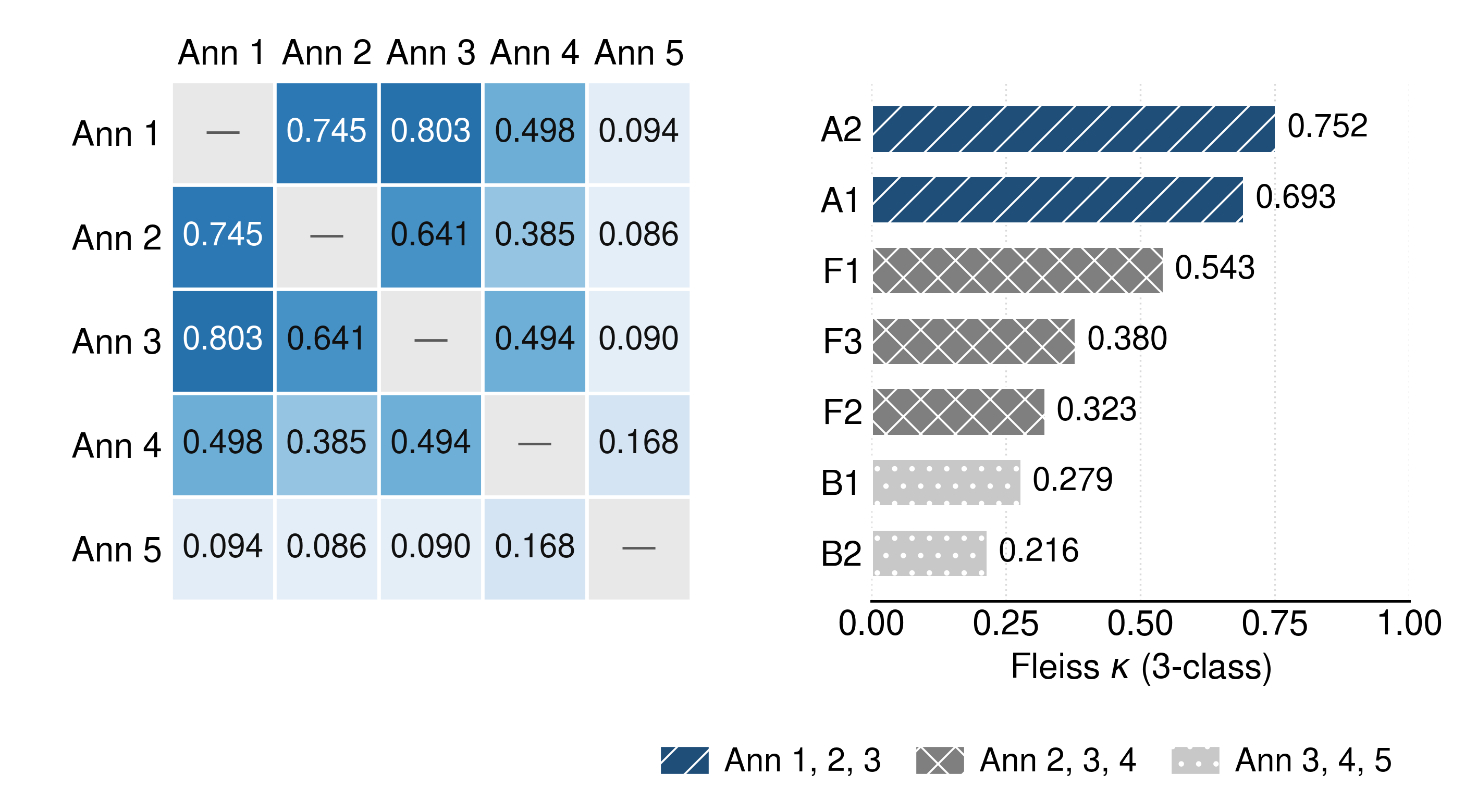}
\caption{Pairwise Cohen's $\kappa$ on the reference set (left; three-class,
$n = 186$) and batch-level Fleiss $\kappa$ by annotator triad (right;
three-class)}
\label{fig:iaa_graphs}
\end{figure}

Figure~\ref{fig:iaa_graphs} reveals that batch-level agreement is primarily driven by annotator composition rather than content. The heatmap shows a clear cluster structure among Ann~1, Ann~2, and Ann~3, who reach strong pairwise Cohen's
$\kappa$~\citep{cohen1960coefficient} up to 0.803, while Ann~4 and Ann~5 show moderate to lower agreement with the rest of the group\footnote{The batch-level Fleiss $\kappa$ values in Figure~\ref{fig:iaa_graphs}
differ from the corresponding Krippendorff's $\alpha$ values in
Table~\ref{tab:iaa} only at the third decimal place. This is expected
rather than redundant: for nominal data with a fixed number of raters
and no missing annotations, the two coefficients differ only in their
chance-correction term, with $\alpha$ applying a small-sample correction
to the expected disagreement, so that
$\alpha \approx \kappa$ as $n$ grows \citep{krippendorff2011computing}. We report
$\kappa$ at batch level for comparability with the pairwise Cohen's
$\kappa$ heatmap, and $\alpha$ in Table~\ref{tab:iaa} for consistency
with the training-phase analysis.}. The batch plot makes the consequence of this structure explicit: batches involving the Ann~1--2--3 triad (A1, A2) reach substantial Fleiss 
$\kappa$~\citep{fleiss1971measuring}, while batches involving Ann~5 (B1, B2) fall to the fair agreement range, pulling the pooled average down. Crucially, the two sets of batches differ not in topic or annotation round but solely in annotator composition, which suggests that some annotator pairs converge more naturally on this task than others. This is not unexpected given the inherent difficulty of enthymeme annotation, and motivates the validator roles described in Section~\ref{annsplits}.

\section{Computational Experiments: Signal vs Noise}
\label{sec:experiments}

In this section, we report a set of preliminary computational experiments conducted on the final annotated dataset. The goal is not to establish a state-of-the-art system for enthymeme detection, but to investigate whether the label variation and genuine annotator disagreement captured during annotation constitutes a meaningful signal for machine learning, rather than noise to be resolved prior to training.

Standard practice in supervised NLP treats annotator disagreement as a problem to be eliminated, through majority vote, adjudication, or annotation protocols designed to maximize inter-annotator agreement. This framing assumes that a single ground truth exists for every instance. In tasks where the target phenomenon is interpretative in nature, however, disagreement may reflect genuine ambiguity rather than annotation error \citep{demarneffe2012veridicality,plank-2022-problem,uma-etal-2021-learning}.

We thus test different training strategies to investigate to what extent genuine annotator disagreement yields signal for enthymeme detection.

\subsection{Models and Training}

 We first establish lexical baselines, then fine-tune DeBERTa-v3-base \citep{he-etal-2023-debertav3}, an ELECTRA-style refinement of DeBERTa \citep{he-etal-2021-deberta} and a strong encoder for short informal texts whose disentangled attention mechanism has shown robust performance on noisy social media data \citep{barbieri-etal-2020-tweeteval}, and finally experiment with soft labels derived from the Dawid-Skene algorithm to explicitly model annotator disagreement in the training data.

\textbf{Lexical Baselines.}
We evaluate two classical baselines using TF-IDF representations: a Linear Support Vector Classifier \citep{cortes-vapnik-1995} and Logistic Regression \citep{cox-1958}. Both models operate on unigram and bigram features and serve to establish whether surface-level lexical patterns alone are sufficient for enthymeme detection.

\textbf{DeBERTa Fine-tuning.}
We fine-tune DeBERTa-v3-base \citep{he-etal-2023-debertav3} on the binary classification task using majority-vote labels derived from our annotations. The model is trained with class-weighted cross-entropy loss to account for class imbalance, and layer-wise learning rate decay to preserve pretrained representations in lower layers.\footnote{DeBERTa-v3-base was fine-tuned for 5 epochs with a batch size of 16, a maximum sequence length of 128 tokens, and a peak learning rate of 2e-5 with linear warmup over 10\% of training steps. Layer-wise learning rate decay was set to 0.9.}

\textbf{Soft Label Training with Dawid-Skene.}
We replace majority-vote labels with per-instance probability distributions estimated via the Dawid-Skene algorithm \citep{dawid-skene-1979}, which infers the latent true label for each instance while accounting for individual annotator profiles. The same DeBERTa architecture is then trained against these soft targets using focal loss \citep{lin-etal-2017-focal}, allowing the model to remain appropriately uncertain on genuinely ambiguous instances rather than committing to a hard decision.

\subsection{Results}

The dataset was split into development and test partitions preserving the global class distribution through stratified sampling. The development set (n=1333) was used for model selection and hyperparameter tuning via 5-fold cross-validation, ensuring that each fold maintains the original class ratio between implicit and non-implicit instances. The held-out test set (n=149) was kept unseen until final evaluation and serves as an unbiased estimate of generalization performance.

\begin{table}[h]
\centering
\caption{Results on development and test sets. P, R, F1 denote precision, recall, and F1-score. DS = Dawid-Skene. Best F1 per partition in \textbf{bold}}
\label{tab:results}
\small
\begin{tabular}{lcccccccc}
\toprule
& & & \multicolumn{3}{c}{\textbf{None}} & \multicolumn{3}{c}{\textbf{Implicit}} \\
\cmidrule(lr){4-6} \cmidrule(lr){7-9}
\textbf{Model} & \textbf{Acc} & \textbf{Mac-F1} & \textbf{P} & \textbf{R} & \textbf{F1} & \textbf{P} & \textbf{R} & \textbf{F1} \\
\midrule
\multicolumn{9}{l}{\textit{Development (5-fold CV, n=1333)}} \\
\midrule
TF-IDF + SVC    & 0.667 & 0.425 & 0.667 & 0.993 & 0.798 & 0.684 & 0.029 & 0.055 \\
TF-IDF + LR     & 0.658 & 0.629 & 0.759 & 0.708 & 0.732 & 0.495 & 0.561 & 0.526 \\
DeBERTa Hard    & 0.694 & 0.676 & 0.805 & 0.706 & 0.752 & 0.543 & 0.671 & 0.600 \\
DeBERTa DS      & 0.718 & \textbf{0.694} & 0.807 & 0.754 & 0.780 & 0.574 & 0.648 & 0.608 \\
\midrule
\multicolumn{9}{l}{\textit{Test (held-out, n=149)}} \\
\midrule
TF-IDF + SVC    & 0.642 & 0.391 & 0.651 & 0.979 & 0.782 & 0.000 & 0.000 & 0.000 \\
TF-IDF + LR     & 0.601 & 0.575 & 0.716 & 0.649 & 0.681 & 0.433 & 0.510 & 0.469 \\
DeBERTa Hard    & 0.669 & 0.647 & 0.773 & 0.701 & 0.735 & 0.517 & 0.608 & 0.559 \\
DeBERTa DS      & 0.689 & \textbf{0.676} & 0.815 & 0.680 & 0.742 & 0.537 & 0.706 & 0.610 \\
\bottomrule
\end{tabular}
\end{table}

Table~\ref{tab:results} reports results across all models on both partitions. The lexical baselines confirm that surface-level features are insufficient for enthymeme detection: TF-IDF with SVC collapses entirely on the test set, predicting the majority class for every instance, while Logistic Regression recovers some implicit cases but remains well below the neural models. DeBERTa fine-tuning brings consistent gains across all metrics. The dev-to-test gap is small and consistent across models, suggesting that the stratified split yields representative partitions and that no model is overfitting to the development folds. Soft label training with Dawid-Skene yields the best performance across both partitions, with gains that are most pronounced on implicit recall.

\subsection{Annotator Disagreement as Signal}

The consistent gains of the soft-label model over its hard-label counterpart are particularly informative given that the two models share the same architecture and data, differing in how annotator disagreement is treated — discarded via majority vote, or preserved as a calibrated probability distribution — along with the training objective that accompanies each labeling strategy. While these preliminary results do not isolate the contribution of disagreement information, they give a clear idea of what preserving label variation can contribute to the task.\footnote{These results should be taken as preliminary evidence rather than a controlled demonstration: assessing their robustness and isolating the specific contribution of annotator disagreement would require additional control experiments beyond the scope of this resource paper, which we leave to future work.}

The gains are most pronounced on implicit recall, rising from 0.608 to 0.706 on the test set. This is where the improvement was expected: implicit enthymemes are the instances where annotators disagree most. Majority vote suppresses these cases toward the none class, while soft-label modeling retains their ambiguity in the training target, allowing the model to learn from borderline instances rather than discarding them as noise.

This finding has an implication for dataset construction. Maximizing inter-annotator agreement would reduce exactly the kind of variation that proves most informative here. Our results suggest that for interpretative tasks of this kind, preserving disagreement rather than resolving it is not only theoretically motivated \citep{plank-2022-problem,uma-etal-2021-learning} but also empirically beneficial.

\section{Discussion}
\label{sec:discussion}

\subsection{Disagreement Patterns in Enthymeme Annotation}
\label{sec:discussion-variation}

Enthymeme annotation is an interpretive task in which disagreement is not incidental but structural. In \citet{pastor2026disagreement}, we distinguish two types of disagreement in enthymeme annotation: \textit{inconsistencies}, arising from task complexity and cognitive load, and \textit{genuine disagreement}, reflecting irreducible interpretive variation rooted in subjective perception and individual inferential processes. This distinction maps onto two linguistic levels — semantic and pragmatic — each with its own sources and implications for annotation design.

\subsubsection{Semantic Level}
At the semantic level, disagreement concerns the identification and segmentation of propositions from which an argument structure can be recovered. For instance, one of the sources of both inconsistent annotation and genuine disagreement here is the presence of \textit{multiple propositions} in a single tweet: when several enthymemes can be reconstructed from the same source, annotators diverge both through cognitive load and through legitimate differences in which reading they consider soundest. The example in Table~\ref{tab:tweet198} illustrates this clearly.

\begin{table}[h]
\centering
\caption{Tweet 198: multiple propositions}
\label{tab:tweet198}
\small
\begin{tabular}{@{} p{1.5cm} l p{7.5cm} @{}}
\toprule
\multicolumn{3}{p{9cm}}{\textit
{\textbf{Tweet 198:} THIS IS AN ABSOLUTE OUTRAGE???? YET THEY NOW WANTING NO STUDIES DONE???? SOMETHING NEED TO BE DONE HERE?? They care less about loss of life?? it's all about SOCIALIST control?? Caught Using False Data To Recommend Kids' COVID Vaccine}} \\
\midrule
\textbf{A1} & \textsc{p1} & {[CDC?]} Caught Using False Data To Recommend Kids' COVID Vaccine \\
& \textsc{p2} & \textcolor{blue}{If {[the CDC]} is caught using false data to recommend Kids' COVID Vaccine, this is an absolute outrage and something must be done} \\
& \textsc{conc.} & THIS IS AN ABSOLUTE OUTRAGE --- SOMETHING NEED TO BE DONE HERE \\
\midrule
\textbf{A2} & \textsc{p1} & Authorities used false data to recommend children's COVID vaccine \\
& \textsc{p2} & They now want no further studies done on those vaccines \\
& \textsc{p3} & \textcolor{blue}{Using manipulated data to push childhood vaccines while suppressing scientific scrutiny serves a control agenda rather than public health} \\
& \textsc{conc.} & This is all about socialist control and is an absolute outrage \\
\midrule
\textbf{A3} & \textsc{p1} & {[CDC]} Recommend Kids' COVID Vaccine \\
& \textsc{p2} & {[CDC]} Caught Using False Data To Recommend Kids' COVID Vaccine \\
& \textsc{conc.} & THIS IS AN ABSOLUTE OUTRAGE???? \\
\bottomrule
\end{tabular}
\end{table}

A1 and A2 both assign an enthymeme label, yet anchor their reconstructions to different propositional subsets of the tweet: A1 focuses on the CDC's use of false data as the trigger for outrage, while A2 constructs a broader implicit claim connecting data manipulation to a political control agenda. A3, faced with the same propositional density, defaults to \textsc{none}. This case illustrates how \textit{multiple propositions} simultaneously occasion inconsistencies through higher cognitive load — captured by the Discrimination and Expressiveness dimensions of \citet{fort-etal-2012-modeling} — and genuine disagreement through the legitimate divergence in which argument each annotator considers soundest.

\subsubsection{Pragmatic Level}

At the pragmatic level, disagreement arises once propositional content has been recovered, as annotators must bridge the gap between what is literally encoded and what the author intended. In \citet{pastor2026disagreement} we identify two illustrative sources of genuine pragmatic disagreement: \textit{commonsense} reasoning and \textit{argument schemes}.
In the commonsense case (Tweet 1829: \textit{``Choose not to get the joke Covid vaccines? Choose to travel freely? Choose WHAT? This? Such a good dictator.''}), A3 recovers the implicit conclusion \textit{``there is no real choice''} via the commonsense premise that a choice requires two viable options, while A1 and A2 assign \textsc{none}. The implicit link is not recoverable from the text alone; it depends entirely on subjective perception of commonsense.

In the argument schemes case (Tweet 3138: \textit{``Vital we continue to crack down on the organized gangs involved in immigration crime [...] Excellent work by the @NCA\_UK.''}), A3 reconstructs an \textit{Argument from Goal} scheme connecting the NCA's work to the broader crackdown objective, while A1 provides a different but equally adequate reconstruction and A2 sees no enthymeme. All three responses are consistent, demonstrating that annotators apply schemes in individual but stable ways — producing genuine variation.

\subsubsection{Summary}
 We find that genuine pragmatic disagreement is structurally distinct from inconsistent annotations: the latter are caused by the inflation of annotation complexity dimensions, in particular Discrimination, Expressiveness, and Context \citep{fort-etal-2012-modeling}, and arise, at the pragmatic level, where disambiguating context is materially recoverable but costly to retrieve; genuine disagreement, by contrast, arises precisely where no such material trace exists, originating instead in the subjective perception and individual inferential processes that annotators bring to the task. In summary, we find more genuine disagreement when annotators reconstruct an interpretation from scratch than when they can depend on potential exterior sources. This variation, as our soft-label experiments suggest, constitutes a meaningful training signal rather than noise to be resolved.

\subsection{Political Content Analysis of Implicit Reconstructions}
\label{sec:discussion-findings}

To characterize the political content that annotators reconstruct as implicit in enthymematic tweets,
we conducted a keyword-based analysis of all 1{,}403 individual implicit annotations
(1{,}082 COVID-19-related, 321 immigration-related), including minority-vote annotations
where individual annotators identified an enthymeme despite no majority agreement.
No NLP classifiers were used; subtopics were identified through regular-expression
keyword clusters applied to the reconstructed text.

Table~\ref{tab:subtopics} reports the frequency of the main political subtopics
recovered from the implicit layer in each domain.
In the COVID-19 domain, mandates and coercion constitute the largest single cluster (14.3\%),
followed closely by the abortion--vaccine hypocrisy frame (12.0\%), concerns about
effectiveness (9.8\%) and rushed development (9.7\%).
In the immigration domain, the three leading clusters are nearly equal in size:
the \emph{illegal/undocumented} framing (14.6\%), UK party-political references
(Labour, UKIP, Brexit, Tories; 14.0\%), and the positive evaluative frame
(\emph{is good/desirable to welcome}; 13.7\%).

\begin{table}[h]
\caption{Main political subtopics recovered from implicit reconstructions (top six per domain)}
\label{tab:subtopics}
\centering\small
\begin{tabular}{lrr}
\toprule
\textbf{Subtopic} & \textbf{N} & \textbf{\%} \\
\midrule
\multicolumn{3}{l}{\textit{COVID-19 (N\,=\,1{,}082)}} \\
Mandates / coercion            & 155 & 14.3 \\
Abortion--vaccine hypocrisy    & 130 & 12.0 \\
Effectiveness / immunity       & 106 &  9.8 \\
Experimental / rushed dev.     & 105 &  9.7 \\
Distrust of pharma / govt      &  94 &  8.7 \\
Bodily autonomy                &  77 &  7.1 \\
\midrule
\multicolumn{3}{l}{\textit{Immigration (N\,=\,321)}} \\
Illegal / undocumented framing        & 47 & 14.6 \\
UK party politics (UKIP/Labour/Brexit)& 45 & 14.0 \\
Good / desirable to welcome           & 44 & 13.7 \\
Points-based / legal immigration      & 36 & 11.2 \\
Mass immigration / open borders       & 28 &  8.7 \\
Border control / sovereignty          & 27 &  8.4 \\
\bottomrule
\end{tabular}
\end{table}

Beyond topic frequency, the two domains differ in the linguistic form of what is left implicit.
Conditional reasoning (\emph{if}) appears in 25.2\% of COVID-19 annotations but only
13.4\% of immigration annotations, while normative verdicts
(\emph{is good / is desirable / is important}) appear in 12.1\% of immigration
annotations and just 0.4\% of COVID-19 annotations.
Collective obligation (\emph{we should / we must / our}) is over ten times more
frequent in immigration (11.8\% vs.\ 1.1\%).

\section{Conclusion and Future Work}
\label{sec:conclusion}

 This paper presented a dataset of 1,482 tweets annotated for the presence of enthymemes by five annotators and two validators, with each tweet receiving at least three original annotations and two validations; each annotator, when identifying an enthymeme, also provides its full argument structure, including the reconstructed implicit component. The resource makes two contributions to existing work. First, it restricts annotation to enthymemes whose implicit content carries a contentious political stance, directly engaging with the persuasive and potentially manipulative function of enthymemes in controversial discourse. Second, it proposes a principled solution to a well-documented challenge: reconstructing what was left unsaid in real-world text risks distorting the author's argument, exposing the analyst to charges of misrepresentation. To address this, the annotation framework grounds reconstruction in Walton's argumentation schemes \citep{walton2008}, which provide a traceable inferential basis for each annotation: the scheme's license is reified as a propositional conditional and injected into the premise set, such that the reconstructed argument classically entails its conclusion. This deductive framing makes the completeness of every reconstruction structurally verifiable, while the scheme annotation constrains what may be reconstructed.

At the same time, the definition is deliberately designed to preserve the interpretive openness that is structurally inherent to the task. Rather than suppressing annotator variation, the framework pinpoints where legitimate interpretive variation can enter: at the level of propositional segmentation, where multiple readings of the same tweet are possible, and at the level of scheme selection as well as sarcasm and commonsense reasoning, where different inferential patterns may each yield an adequate reconstruction. The complexity analysis across the six dimensions of \citet{fort-etal-2012-modeling} provides a basis for distinguishing inconsistencies, attributable to the cognitive load of the task, from genuine disagreement, which reflects irreducible differences in individual annotator perception and inferential reconstruction.

Beyond these specific findings, the resource makes a broader contribution at the intersection of annotation methodology and computational argumentation. The complexity analysis can serve as a diagnostic tool for semantic interpretation tasks more generally: profiling a task along the six dimensions makes it possible to anticipate, before a campaign begins, where annotation quality will degrade under cognitive load and where variation should instead be expected and preserved. The computational experiments provide preliminary empirical evidence that preserving such genuine disagreement through soft-label training improves model performance, in line with methodological arguments in the annotation literature. The data made available here, including full propositional decompositions and argument scheme annotations across five annotators, further opens the resource to research on how implicit meaning is reconstructed differently across individuals and backgrounds. Finally, enthymeme detection of the kind studied here points toward practical downstream applications: a pipeline that identifies enthymematic arguments, reconstructs their implicit premises, and submits those premises to automated fact-checking could provide a systematic means of tracking whether a discourse is structurally relying on false or unverifiable claims.

\section{Ethical Considerations}
\label{sec:ethics}

\subsection{Source Data, Redistribution, and Licensing}

The tweets in our resource originate from the dataset of \citet{Flaccavento2025}, which was released with links stripped and personally identifiable information removed; remaining user mentions refer to public figures and institutions. X's (formerly Twitter) terms of service restrict the redistribution of tweet content, and the customary workaround is to release only tweet IDs for rehydration. This route is not available here: the source dataset was released in anonymized form without tweet or author IDs, so the texts cannot be rehydrated --- nor, conversely, can they be traced back to the accounts that produced them. We therefore redistribute the anonymized texts themselves, following the precedent set by the source dataset, and consider that the removal of all identifiers substantially mitigates the privacy risks that the redistribution restrictions are designed to address. The resource is released under the license stated in the Data Availability statement.

The dataset contains unfiltered profanity, offensive language, and misinformation. It is released for research on argumentation, manipulation, and misinformation; it is not intended for training systems that generate persuasive content, and the implicit reconstructions it contains should not be read as endorsements of the reconstructed claims.

\subsection{Annotation Campaign}

All annotators participated either as volunteers or under compensated agreements (Section~\ref{sec:process}). They were informed in advance of the nature and purpose of the task and of the presence of offensive and controversial content, and could interrupt their participation at any time. All participants gave their consent to the use of their annotations in this resource and to being named in the Acknowledgements. No personal data beyond the annotations themselves was collected during the campaign.

\begin{acknowledgements}
This work was produced as part of the HYBRIDS project, a Marie Skło\-dow\-ska-Curie Doctoral Network funded by the European Union under grant no. 101073351 and the UK Research and Innovation (UKRI) Horizon Funding Guarantee, and the AI-CODE project, funded under the European Union's Horizon Europe research and innovation programme grant agreement no. 101135437. We also acknowledge the annotators: Nienke Gelderland, Afrooz Bolboli, Jarno Van Arkel, Peter Beinema, Ulas Askin, Huixin Lan and Pascale Mannone.
\end{acknowledgements}

\section*{Declarations}

\paragraph{Funding.} This work was supported by the HYBRIDS project, a Marie Sk{\l}o\-dow\-ska-Curie Doctoral Network funded by the European Union under grant no.\ 101073351 and the UK Research and Innovation (UKRI) Horizon Funding Guarantee, and by the AI-CODE project, funded under the European Union's Horizon Europe research and innovation programme, grant agreement no.\ 101135437.

\paragraph{Conflict of interest.} The authors have no competing interests to declare that are relevant to the content of this article.

\paragraph{Ethics approval and consent to participate.} The annotation campaign involved no collection of personal data beyond the annotations themselves; all participants gave informed consent to participate and to be named in the Acknowledgements (see Section~\ref{sec:ethics}).

\paragraph{Consent for publication.} All participants in the annotation campaign
consented to being named in the Acknowledgements. The tweets reproduced in this
article are anonymised and carry no personally identifying information
(see Section~\ref{sec:ethics}).

\paragraph{Data availability.} The dataset --- all individual annotations,
propositional decompositions, and argument scheme reconstructions --- together with
the annotation guidelines (Section~\ref{sec:guidelines}), which include a screenshot
and walkthrough of the annotation tool, is publicly available at \repoURL{} under a
\datasetLicence{} licence.

\paragraph{Code availability.} The annotation tool is available in the same
repository, \repoURL{}, under the same licence.

\paragraph{Author contributions.} M.P.\ designed the study, developed the
annotation framework and guidelines, built the annotation tool, ran the annotation
campaign, carried out the experiments and analysis, and wrote the manuscript.
N.O.\ contributed to the design and organisation of the annotation campaign and to
the writing and revision of the manuscript. Both authors read and approved the
final manuscript.

\clearpage
\printbibliography

\clearpage
\appendix
\section{Appendix}

\begin{figure}[!ht]
\centering
\begin{tikzpicture}[x=2.6cm, y=1.05cm, font=\small]
  \definecolor{ctxa}{gray}{0.90}
  \definecolor{ctxb}{gray}{0.72}
  \definecolor{ctxc}{gray}{0.44}
  \definecolor{ctxd}{gray}{0.26}
  \fill[ctxa] (0,0) -- (1,0) -- (0,1) -- cycle;
  \fill[ctxb] (1,0) -- (2,0) -- (0,2) -- (0,1) -- cycle;
  \fill[ctxc] (2,0) -- (3,0) -- (0,3) -- (0,2) -- cycle;
  \fill[ctxd] (3,0) -- (3,3) -- (0,3) -- cycle;
  \foreach \k in {1,2,3}{\draw[white, line width=0.9pt] (0,\k) -- (\k,0);}
  \draw[gray!55, line width=0.3pt] (0,0) rectangle (3,3);

  \node[font=\footnotesize\bfseries, text=black] at (0.30,0.24) {0.25};
  \node[font=\footnotesize\bfseries, text=black] at (0.58,0.86) {0.50};
  \node[font=\footnotesize\bfseries, text=white] at (1.28,1.18) {0.75};
  \node[font=\footnotesize\bfseries, text=white] at (2.05,2.10) {1};

  \draw[-{Latex[length=4pt]}, line width=0.6pt] (0,0) -- (0,3.45);
  \draw[-{Latex[length=4pt]}, line width=0.6pt] (0,0) -- (3.45,0);

  \foreach \y/\l in {0/0, 1/Sentence, 2/Paragraph, 3/Whole text}{%
    \draw[line width=0.6pt] (-0.07,\y) -- (0,\y);
    \node[anchor=east, font=\footnotesize, xshift=-3pt] at (0,\y) {\l};}
  \foreach \x/\l in {1/{Annotation\\guide}, 2/{Identified external\\sources}}{%
    \draw[line width=0.6pt] (\x,-0.07) -- (\x,0);
    \node[anchor=north, font=\scriptsize, align=center, yshift=-3pt]
      at (\x,0) {\l};}
  \draw[line width=0.6pt] (3,-0.07) -- (3,0);
  \node[anchor=north, font=\scriptsize\itshape\bfseries, align=center, yshift=-3pt]
    at (3,0) {New sources\\to identify};

  \draw[dashed, line width=0.7pt] (0,1.5) -- (3,1.5);
  \node[draw=black, fill=white, line width=0.9pt, circle, inner sep=1.7pt]
    at (3,1.5) {};
  \node[anchor=east, font=\footnotesize\itshape\bfseries, xshift=-3pt]
    at (0,1.5) {Tweet};

  \node[anchor=south west, font=\footnotesize\bfseries, align=left]
    at (-0.24,3.5) {Size of co-text};
  \node[anchor=north, font=\footnotesize\bfseries, align=center]
    at (1.5,-0.95) {Accessibility of knowledge sources};
\end{tikzpicture}
\caption{Co-text weight, after \citet{fort2012ressources}. The vertical axis gives the
size of the text span to be taken into account, the horizontal axis the accessibility of
the sources to be consulted; the two qualitative scales combine into a single discrete
scale from 0 to 1}
\label{fig:context-weight}
\end{figure}

{\small
\begin{xltabular}{\textwidth}{>{\bfseries}l X}
\caption{Controversial topics and their associated implicit stances for the
Immigration and Covid-19 domains}
\label{tab:semantic-space}\\
\toprule
Topic & Implicit controversial stance \\
\midrule
\endfirsthead
\caption[]{Controversial topics and their associated implicit stances
\textit{(continued)}}\\
\toprule
Topic & Implicit controversial stance \\
\midrule
\endhead
\midrule
\multicolumn{2}{r}{\textit{continued on next page}} \\
\endfoot
\bottomrule
\endlastfoot

\multicolumn{2}{l}{\textit{\textbf{Immigration}}} \\
\midrule

Refugee integration      & Migrants drive down working-class wages. \\
and employment           & Immigration shows the UK is an attractive place to work and study. \\
                           & Supporting refugee employment and integration is desirable. \\
                         & Open borders and expanded migrant rights are harmful to the working class. \\
                         & Refugees are a potential danger and threat. \\
                         & Radical left Democrats are as responsible for taking in immigrants. \\
                         & The idea that immigrants strengthen the destination country is false. \\
                         & Welsh children should come before migrants. \\
                         & Taxpayer money should not fund college for illegal immigrants. \\
                         
\midrule

Border security          & Maintaining secure borders is important.\\
and immigration control  & Immigration, border control, and repatriation of illegal immigrants should be closely monitored. \\
                         & Uncontrolled immigration is undesirable. \\
                         & Deterring illegal immigration is firm but fair. \\
                         & Illegal migrants should not enter or remain in Europe or the UK. \\
                         & Migrant smugglers can save lives. \\
                         & Migrant smugglers reduce inequality despite their exploitative role. \\
                         & Twitter censors people who report the ``truth'' about illegal immigration. \\
\midrule

Immigration levels       & Immigration is too high, and Brexit is seen as the solution. \\
and Brexit               & Brexit will end mass migration. \\
                         & Capitalists want to replace Europeans with non-European migrants. \\
\midrule

Immigration policy       & The Labour party prioritizes votes over immigration control. \\
                         & The Labour party is inconsistent and no longer cares about the working class. \\
                         & Leaders support open borders to protect their leadership position. \\
                         & Supporting UKIP is the way to stop illegal immigration. \\
                         & Immigration rules should treat UK and EU citizens equally and fairly. \\
                         & Traditional Labour voters will not accept open-door migration. \\
\midrule

\multicolumn{2}{l}{\textit{\textbf{Covid-19}}} \\
\midrule

Vaccination mandates     & Vaccine mandates are wrong and unjustified. \\
                         & Democrats are disconnected from public opinion on COVID-19 mandate policies. \\
                         & Vaccines increase the risk of hospitalization. \\
                         & Those who oppose vaccination or mask mandates are not genuinely pro-life. \\
                         & Liberals are hypocritical in their stance on bodily autonomy during COVID-19. \\
                         & Supporting vaccine mandates contradicts the principle of `my body, my choice'. \\
                         & COVID-19 vaccine trials are scientifically unreliable. \\
                         & People who say they were forced to get vaccinated are ignorant. \\
\midrule

Vaccine deployment       & Pharma manipulates vaccine data. \\
and institutional        & Vaccine efficacy claims based on hope are deceptive. \\
opportunism              & Vaccines are prepared and deployed opportunistically to coincide with disease outbreaks. \\
                         & Public officials acted misleadingly regarding the effectiveness of COVID-19 vaccines. \\
                         & Vaccine-injured people are unjustly silenced and should be allowed to speak about vaccine harms. \\
                         & Twitter censors information critical of COVID-19 vaccines to protect financial interests. \\
                         & The vaccine industry profits from recurring vaccines. \\
\end{xltabular}
}

\end{document}